\documentclass[journal]{IEEEtran}
\ifCLASSINFOpdf
  \usepackage[pdftex]{graphicx}
\else
  \usepackage[dvips]{graphicx}
\fi
\usepackage{amsmath}
\usepackage{amssymb}
\usepackage{bm}
\usepackage[ruled,linesnumbered]{algorithm2e}
\usepackage{algpseudocode}

\usepackage{array}
\newcolumntype{P}[1]{>{\centering\arraybackslash}p{#1}}
\usepackage{booktabs}
\usepackage{caption}

\begin{document}
\bstctlcite{IEEEexample:BSTcontrol}

%
\title{Learning to Synthesize Compatible Fashion Items Using Semantic Alignment and Collocation Classification: An Outfit Generation Framework}
%
%
%

\author{Dongliang~Zhou, Haijun~Zhang,
        Kai~Yang, Linlin~Liu, Han~Yan, \\ Xiaofei~Xu, Zhao~Zhang,\ and Shuicheng~Yan
\thanks{This work was supported in part by the National Natural Science Foundation of China under Grant no. 61972112 and no. 61832004, the Guangdong Basic and Applied Basic Research Foundation under Grant no. 2021B1515020088, the Shenzhen Science and Technology Program under Grant no. JCYJ20210324131203009, and the HITSZ-J\&A Joint Laboratory of Digital Design and Intelligent Fabrication under Grant no. HITSZ-J\&A-2021A01.}
\thanks{D. Zhou, H. Zhang, K. Yang, L. Liu, H. Yan and X. Xu are with the Department
of Computer Science, Harbin Institute of Technology,
Shenzhen, 518055 China; Z. Zhang is with the Department of Computer Science, Hefei University of Technology, Hefei, 230009 China; S. Yan is with Sea AI Lab (SAIL), 119077 Singapore. Corresponding author: Haijun Zhang, e-mail: hjzhang@hit.edu.cn.}
}

\maketitle

\begin{abstract}
The field of fashion compatibility learning has attracted great attention from both the academic and industrial communities in recent years. Many studies have been carried out for fashion compatibility prediction, collocated outfit recommendation, artificial intelligence (AI)-enabled compatible fashion design, and related topics. In particular, AI-enabled compatible fashion design can be used to synthesize compatible fashion items or outfits in order to improve the design experience for designers or the efficacy of recommendations for customers. However, previous generative models for collocated fashion synthesis have generally focused on the image-to-image translation between fashion items of upper and lower clothing. In this paper, we propose a novel outfit generation framework, i.e., \textit{OutfitGAN}, with the aim of synthesizing a set of complementary items to compose an entire outfit, given one extant fashion item and reference masks of target synthesized items. OutfitGAN includes a semantic alignment module, which is responsible for characterizing the mapping correspondence between the existing fashion items and the synthesized ones, to improve the quality of the synthesized images, and a collocation classification module, which is used to improve the compatibility of a synthesized outfit. In order to evaluate the performance of our proposed models, we built a large-scale dataset consisting of 20,000 fashion outfits. Extensive experimental results on this dataset show that our OutfitGAN can synthesize photo-realistic outfits and outperform state-of-the-art methods in terms of similarity, authenticity and compatibility measurements.
\end{abstract}


%
\IEEEpeerreviewmaketitle

\section{Introduction}
%
%
%
%
\IEEEPARstart{I}{n} recent years, the fashion sector has undergone a proliferation in economic terms. According to a business report\footnote{https://www.statista.com/outlook/244/100/fashion/worldwide} from Statista.com, the economy is expected to maintain an approximate annual growth rate of 7.2\% in the future. A Mckinsey.com business report\footnote{https://www.mckinsey.com/industries/retail/our-insights/the-state-of-fashion-2020-navigating-uncertainty} recommends that fashion sellers should pay more attention to cutting-edge techniques, as these offer lucrative opportunities. The key to improving revenue for sellers lies in fashion designers creating more attractive fashion items or outfits for customers. In a traditional design process, however, designers rely on their own creative senses, which may involve subjectivity and uncertainty. With the advent of artificial intelligence (AI) and the era of big data, AI-enabled fashion design has become possible. Fashion designers can create preliminary designs more effectively by relying on machine learning based on numerous extant collocated outfits shared by social media users. The rules of compatibility hidden in these collocated outfits can be learned by a machine learning model to produce new fashion items.
\begin{figure}[!t]
    \centering
    \setlength{\abovecaptionskip}{0.cm}
    \includegraphics[width=0.5\textwidth]{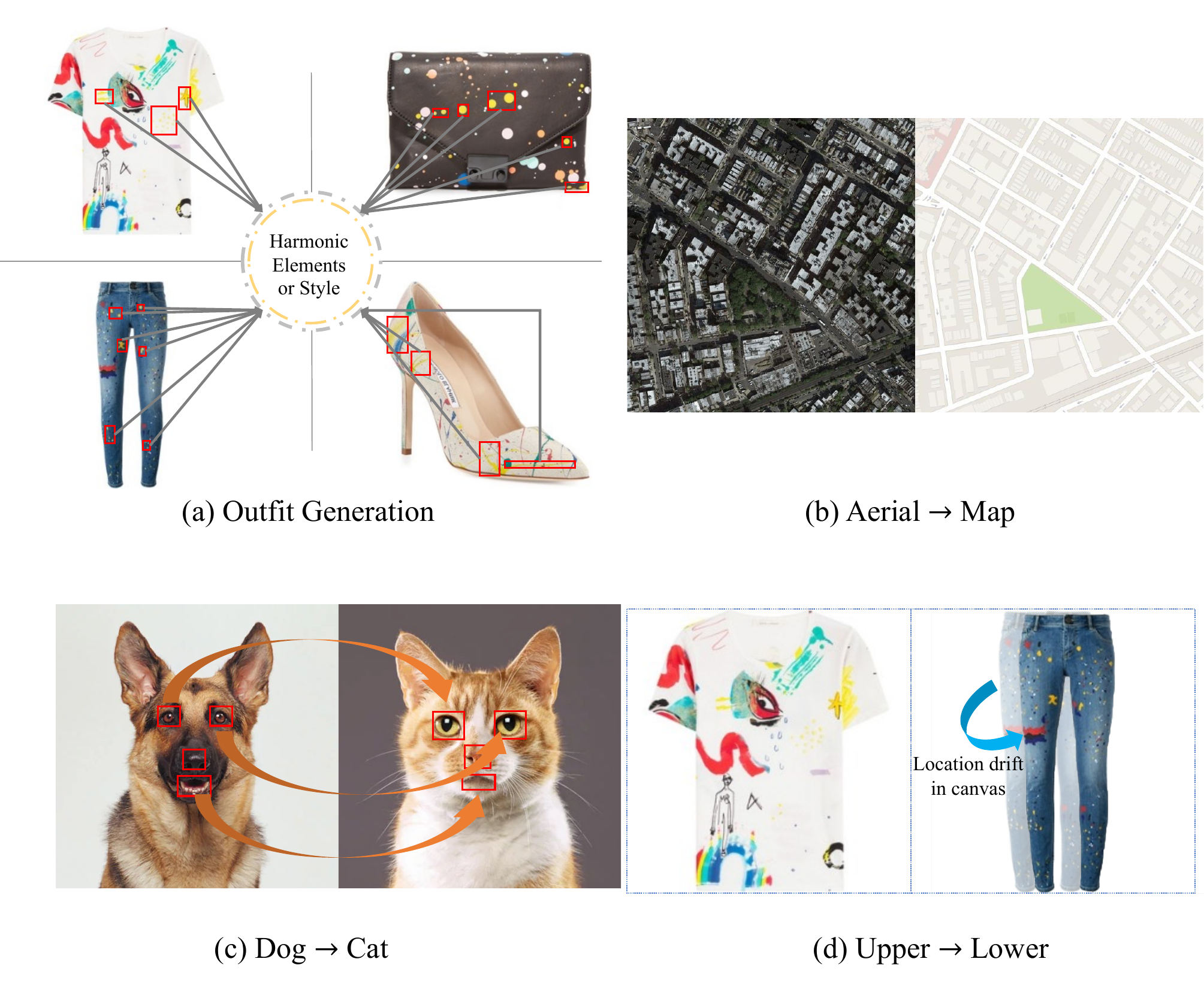}
    \caption{General image-to-image translation and fashion outfit generation.}
    \label{introduction_of_paper}
\end{figure}
In particular, generative adversarial networks (GANs) \cite{NIPS2014_gan} can assist fashion designers in synthesizing visually plausible images of new fashion items based on extant fashion items. This can be approached as a direct image-to-image translation task in which GANs are fed with pair-wise image data containing extant fashion items and corresponding compatible items during training. The use of GAN-based models has been widely explored in the field of image synthesis. For general image-to-image translation, Isola \textit{et al.} \cite{pix2pix2017} proposed a Pix2Pix framework for the synthesis of new images based on extant ones, by adopting a GAN loss and an L1 loss to improve the photo-realism of images with high authenticity. In a later study, Wang \textit{et al.} \cite{wang2018pix2pixHD} improved the Pix2Pix framework in a coarse-to-fine manner. In addition to ground-truth images, several unsupervised image-to-image translation methods such as CycleGAN \cite{CycleGAN2017}, MUNIT \cite{huang2018munit}, DRIT++ \cite{DRIT_plus} and StarGAN-v2 \cite{choi2020starganv2} were explored for image-to-image translation using images from two arbitrary domains, without the explicit use of one-to-one correspondent mapping. In the field of fashion synthesis, Liu \textit{et al.} \cite{attribute_gan} first proposed an Attribute-GAN model, which addressed the task of translation between upper clothing and lower clothing items by considering the latent compatibility between them. They added clothing attributes to guide the process of image generation. In a later study, they \cite{multi_dis_fashion} extended their collocation discriminator and attribute discriminator framework to form a multi-discriminator framework. Yu \textit{et al.} \cite{Yu_2019_ICCV} used the user's personal preferences to improve the generation quality of clothing images for personalized recommendation. All of the aforementioned works in fashion synthesis focused on image-to-image translation between upper and lower clothing, and rarely considered the generation of a whole outfit. 

By taking advantage of the power of GANs, the objective of this research is to explore the issue of how to synthesize an entire compatible outfit, based on extant fashion items with certain alternative reference information, e.g., outline mask information of target fashion items. To accomplish this, we consider a specific scenario: given a particular fashion item, a user may expect to have other collocated fashion items with desirable reference masks in his or her mind, containing outline information on compatible items. Our model aims to generate compatible fashion items to compose an outfit, conditioned on a particular fashion item and the reference masks of other items in the same outfit. The aim of outfit generation is thus to translate harmonic elements or styles from extant fashion garments to synthesized compatible items. As shown in Fig. \ref{introduction_of_paper}(a), each fashion item in an outfit shares harmonious elements or styles with the other items, to maintain the compatibility. For example, there may be many patches of yellow or elements of the same color co-occurring in an outfit. Meanwhile, in the task of outfit generation, each type of fashion item in an outfit has its own unique outline and style. More specifically, compared with general image-to-image translation methods, our research addresses this problem from the following three perspectives. (i) As shown in Fig.~\ref{introduction_of_paper}(b), traditional supervised image-to-image translation methods such as Pix2Pix \cite{pix2pix2017} or Pix2PixHD \cite{wang2018pix2pixHD} need pixel-to-pixel correspondence between an input image and its output image, such as in aerial-to-map translation. For these translation tasks, researchers usually adopt convolutional neural network (CNN)-based generators, which can learn only local patch features of an input image rather than the global features \cite{wang2018non}. However, outfit generation has no apparent pixel-to-pixel alignment between extant fashion items and synthesized ones. (ii) Unsupervised image-to-image translation methods such as CycleGAN \cite{CycleGAN2017} or MUNIT \cite{huang2018munit} learn a mapping function between two domains. For example, as shown in Fig. \ref{introduction_of_paper}(c), objects need a latent semantic alignment rather than a precise pixel-to-pixel correspondence, in the same way as eye or mouth mapping in dog-to-cat translation. In contrast, in the process of outfit generation, when an extant upper clothing item has a red flower on the top, the target shoes may have a corresponding element or style at the bottom rather than the top of the shoes. This suggests that the mapping relationship between extant fashion items and target items is non-local in space, and the model therefore needs to learn the global feature mapping relationship during training. (iii) In outfit generation, when the target images undergo minor changes in location or size, users may not observe these slight spatial changes. For example, for a given upper clothing image, the target lower clothing image may drift slightly on the canvas, as shown in Fig. \ref{introduction_of_paper}(d). Although this minor change cannot be immediately observed, it is very important in terms of image-to-image translation, since general image-to-image translation methods with paired images usually supervise the generation process by adopting losses such as L1, L2 \cite{pix2pix2017}, and perceptual losses \cite{ledig2017photo}, which require a precise spatial alignment between the synthesized and target images.

To address the above issues, we propose a collocated fashion outfit generation framework called OutfitGAN, which adopts a semantic alignment module (SAM) and a collocation classification module (CCM) to guide the process of outfit generation. The SAM fuses a given fashion item with the reference masks of other compatible items, to align the extracted features from the extant fashion item based on the reference masks. Our usage of reference masks is largely inspired by \cite{fashion_gan} and \cite{Han_2018_viton}, in which researchers used the key points of the human body to guide the fashion synthesis. Specifically, our OutfitGAN introduces an SAM to improve the quality of synthesized images and provide an explanation of the outfit generation process explicitly. The development of the SAM was motivated by the fact that items mostly convey corresponding areas or style from the same outfit, according to our observations from the collected outfits shown in Fig. \ref{introduction_of_paper}(a), where we see that many compatible outfits contain elements or styles corresponding to other items in the same outfit. To ensure the compatibility of a synthesized outfit, we also develop a CCM based on bidirectional long short-term memory (Bi-LSTM) \cite{bi_lstm} to model the compatibility between items. In order to examine the performance of our proposed OutfitGAN, we constructed a large-scale dataset containing 20,000 outfits, each of which was composed of four types of item: upper clothing, bags, lower clothing, and shoes. The results of an extensive set of experiments demonstrate the effectiveness of our proposed framework with respect to various evaluation metrics, in comparison with several state-of-the-art methods. The main contributions of this research can be summarized as follows:
\begin{itemize}
\item To the best of our knowledge, this is the first work to synthesize fashion items based on extant ones in order to create a compatible outfit. The overall framework of our OutfitGAN includes an outfit generator, an outfit discriminator, and a CCM. The results of our experiments indicate that our proposed framework is capable of synthesizing photo-realistic outfit images that are compatible with a given fashion item. 
\item We propose an SAM to improve the quality of the synthesized outfit and characterize the correspondence between a given fashion item and the synthesized items. This module uses two branches based on CNNs to extract the features of a given fashion item and the reference mask of a target item, respectively; a correspondence layer to calculate the spatial relationship matrix with respect to the features of the given fashion item and the reference mask; and an alignment layer to semantically align the features of the given item with the reference mask.
\item We propose a CCM that uses Bi-LSTM to improve the compatibility of the synthesized outfit. This module is primarily applied to guide the compatibility of the synthesized fashion items. It uses a pre-trained CNN to extract the features of the fashion items and two directional LSTMs for compatibility guidance from different directions, from the perspective of human vision.
\end{itemize}
The remainder of this article is organized as follows. Section \ref{related_work} briefly reviews works related to fashion outfit synthesis. Section \ref{method} describes the overall framework of our OutfitGAN framework and the associated details of the implementation. In Section \ref{exp}, we conduct extensive experiments to validate the performance of our model. Section \ref{cnc} concludes the paper and suggests directions for future work.

\section{Related Work}
\label{related_work}
This research falls into the field of fashion learning, which has a large existing body of literature. In this section, we review related works on image-to-image translation, fashion compatibility learning, and fashion synthesis. We also highlight the features of this research in comparison to those of prior works.

\textbf{Image-to-Image Translation.} This is an important task in computer vision. A model takes an image as input and learns a conditional distribution of the corresponding image with a mapping function. There are many applications for this task, such as image colorization \cite{pix2pix2017}, image style transfer \cite{Gatys_2016_CVPR}, super-resolution \cite{ledig2017photo}, and virtual try-on \cite{Han_2018_viton,tileimage}. Numerous previous studies have suggested that GANs \cite{NIPS2014_gan} are capable of producing realistic synthesized images via image-to-image translation. Existing GAN-based translation methods can be roughly divided into two categories: supervised and unsupervised approaches. Using a supervised method, Isola \textit{et al.} \cite{pix2pix2017} proposed a Pix2Pix translation framework to alleviate blurring in this task. Later, Wang \textit{et al.} \cite{wang2018pix2pixHD} introduced an improved Pix2Pix model with the aim of achieving more stable and realistic image generation in a coarse-to-fine manner. Using an unsupervised method, Zhu \textit{et al.} \cite{CycleGAN2017} proposed a cycle consistency loss to handle a lack of paired images. Subsequently, Huang \textit{et al.} \cite{huang2018munit} addressed the latent space of image samples using a composition of style and content code, and used two separate encoders to disentangle these components. Lee \textit{et al.} \cite{DRIT_plus} also disentangled the latent space into a shared content space and an attribute space for each domain. In a later study, Choi \textit{et al.} \cite{choi2020starganv2} extended the concepts of style code and content code, employing a multi-layer perceptron (MLP) to synthesize a diverse range of style codes and injecting them into a decoder to synthesize various images.

\textbf{Fashion Compatibility Learning.} With the increasing popularity of online stores, fashion recommendation is now playing an essential role in online retail. Fashion compatibility learning is an important aspect of fashion recommendation, and researchers have adopted metric learning to predict compatibility. Each fashion item in the same outfit is firstly embedded into a shared space, and the compatibility between items is then evaluated based on the distance between them. A shorter distance or a higher similarity indicates better compatibility, and vice versa. To measure the compatibility between items, McAuley \textit{et al.} \cite{mcauley2015image} proposed a method for comparing the distance between the features extracted by a pre-trained CNN. Veit \textit{et al.} \cite{veit2015learning} then used a SiameseNet to extract visual features to compare the distance between items. These methods regarded the different types of fashion items as the same, and handled them in an embedding space. In order to keep different categories of fashion items with different mappings into embeddings, Vasileva \textit{et al.} \cite{vasileva2018learning} tackled this problem by learning the similarity and compatibility simultaneously, in different spaces, for each pair of item categories. Another inspired idea was to regard the fashion items in the outfit as a sequence from the perspective of human vision. Han \textit{et al.} \cite{han2017learning} adopted Bi-LSTM to learn the compatibility of an outfit in the form of a sequence. The other mainstream idea that has emerged is the use of graph-based networks to address the issue of compatibility, and these methods have attracted the attention of several researchers. In particular, Cui \textit{et al.} \cite{cui2019dressing} and Li \textit{et al.} \cite{li2020hierarchical} employed graph convolutional networks to model the compatibility problem. In this task, fashion compatibility is a crucially important perspective for generating an outfit. In our OutfitGAN, we use Bi-LSTM in our implementation of collocation classification in order to guide the compatibility of the generated items. 

\textbf{Fashion Synthesis.} Due to the ever-increasing demand for fashion applications, fashion synthesis has started to become an important aspect of the field of computer vision \cite{2020arXiv200313988C}. Fashion synthesis includes virtual try-on, pose transformation and the synthesis of compatible fashion items. In the field of virtual try-on, Han \textit{et al.} \cite{Han_2018_viton} employed a thin plate spline (TPS) and a GAN to synthesize new images, given images of the user's body and the target clothing. Subsequently, a new model called characteristic-preserving image-based virtual try-on network (CP-VTON) \cite{cp_vton} was proposed, which included a geometric matching module that could improve the spatial deformation in comparison to TPS. Zhu \textit{et al.} \cite{fashion_gan} proposed FashionGAN to synthesize clothes on a wearer while maintaining consistency with a text description. In addition to virtual try-on, pose transformation is also an important task in fashion synthesis. A model takes a reference image as input and a target pose based on the key points of the human body, and aims to synthesize a pose-guided image of the person while retaining the personal information of the reference image. A network called $\mathrm{PG}^2$ \cite{ma2017pose} was the first to use a two-stage model to address the problem. Later, Siarohin \textit{et al.} \cite{siarohin2018deformable} transformed the high-level features for each part of human body using a technique called deformable skipping. Recently, researchers have turned their attention to the generation of fashion items. In particular, Liu \textit{et al.} \cite{attribute_gan} proposed a network for image-to-image translation between upper and lower clothing using an attribute-based GAN. They extended their model to a more general GAN framework with multiple discriminators by considering rich text descriptions of upper and lower clothing images \cite{multi_dis_fashion}. Yu \textit{et al.} \cite{Yu_2019_ICCV} then exploited a matrix of the user's personal preferences to improve the quality of image generation. Unlike the works in \cite{attribute_gan}, \cite{multi_dis_fashion} and \cite{Yu_2019_ICCV}, we concentrate in this paper on generating an outfit that consists of several compatible fashion items.

\textbf{Features of Our Model:} Several studies have focused on outfit generation using image-to-image translation \cite{pix2pix2017,wang2018pix2pixHD,CycleGAN2017,huang2018munit,DRIT_plus,choi2020starganv2} and compatibility learning \cite{mcauley2015image,veit2015learning,vasileva2018learning,han2017learning,cui2019dressing,li2020hierarchical} for fashion synthesis \cite{attribute_gan,multi_dis_fashion,Yu_2019_ICCV}. Initially, supervised \cite{pix2pix2017,wang2018pix2pixHD} or unsupervised image-to-image translation methods \cite{CycleGAN2017,huang2018munit,DRIT_plus,choi2020starganv2} were used with CNN-based generators to carry out image translation from input images to output images, with or without supervised paired images. However, a CNN-based generator is only able to learn local neighborhood relationships, and is unable to learn the long-range dependences between the input and output images \cite{wang2018non}. Our outfit generation scheme aims to translate harmonic elements and styles while maintaining their compatibility. In particular, our approach characterizes the long-range dependences between the extant fashion items and the synthesized ones. Unlike the general methods described above, our proposed model is capable of accomplishing cross-domain image translation, in which the images may have no pixel-wise alignment but do have a corresponding spatial alignment mapping for the long-range dependences between the input and output images. In particular, our proposed model uses an SAM which aligns the features of the extant fashion items to those of the target items. In contrast, existing fashion compatibility learning methods \cite{mcauley2015image,veit2015learning,vasileva2018learning,han2017learning,cui2019dressing,li2020hierarchical} are used to predict outfit compatibility and give outfit recommendations for an extant fashion database with discriminative models, and rarely consider the synthesis of new compatible outfits based on extant fashion items. Our proposed model synthesizes compatible fashion items based on extant ones, using a generative model. Finally, although several fashion synthesis methods \cite{attribute_gan,multi_dis_fashion,Yu_2019_ICCV} have been used to synthesize complementary fashion items based on extant ones, these methods only carry out image translation between upper and lower clothing, and cannot synthesize an entire outfit. Our proposed model uses multiple generators to synthesize suitable fashion items for the generation of entire outfits. In addition, a CCM is proposed to supervise the compatibility of the synthesized outfit during the generation process.

\section{OutfitGAN for the Generation of Multiple Fashion Items}
\label{method}
In this section, we first formulate our research problem and give the descriptions and definitions needed for outfit generation. We then present the entire OutfitGAN framework. Finally, the implementation details of our proposed models are discussed.
\subsection{Problem Formulation}
In general, previous fashion compatibility learning methods \cite{mcauley2015image,veit2015learning,vasileva2018learning,han2017learning,cui2019dressing,li2020hierarchical} have focused on discriminating the collocation given a set of fashion items. In contrast, generative models allow us to synthesize a entirely new outfit as well as maintaining the collocation for compatibility learning. In this work, we focus on synthesizing a set of compatible items based on a given fashion item in order to compose a complete outfit. Formally, let $\mathcal{O} =[\mathcal{O} _1,\cdots ,\mathcal{O} _i,\cdots ,\mathcal{O} _N]
$ denote an outfit, where $\mathcal{O} _i$ is the  $i$-th fashion item in the outfit arranged in a fixed order based on its categories, i.e., from top to bottom according to perspective of human vision, e.g., [upper clothing, bag, lower clothing, shoes].  $N$ represents the number of fashion items in an outfit. In addition, each fashion item $\mathcal{O} _i\in \mathcal{O}$ is associated with a mask that indicates the outline of  $\mathcal{O} _i$. For each $\mathcal{O} _i$, let $Mask_i$ be the corresponding mask of $\mathcal{O} _i$. Our task is to synthesize a complementary outfit set $\widetilde{\mathcal{O} }$  for a user based on a given fashion item $\mathcal{O} _k$  and reference masks  $[Mask_1,\cdots ,Mask_{k-1},Mask_{k+1},\cdots ,Mask_N]$, which represent the user's rough idea of the outlines of the newly synthesized outfit items. Here, the reference masks may be given by the user, selected by the user from a candidate dataset containing various outlines of fashion items, or produced automatically by a pre-trained generative model.
\subsection{OutfitGAN}
\begin{figure*}[t]
    \centering
    \setlength{\abovecaptionskip}{0.cm}
    \includegraphics[width=1\textwidth]{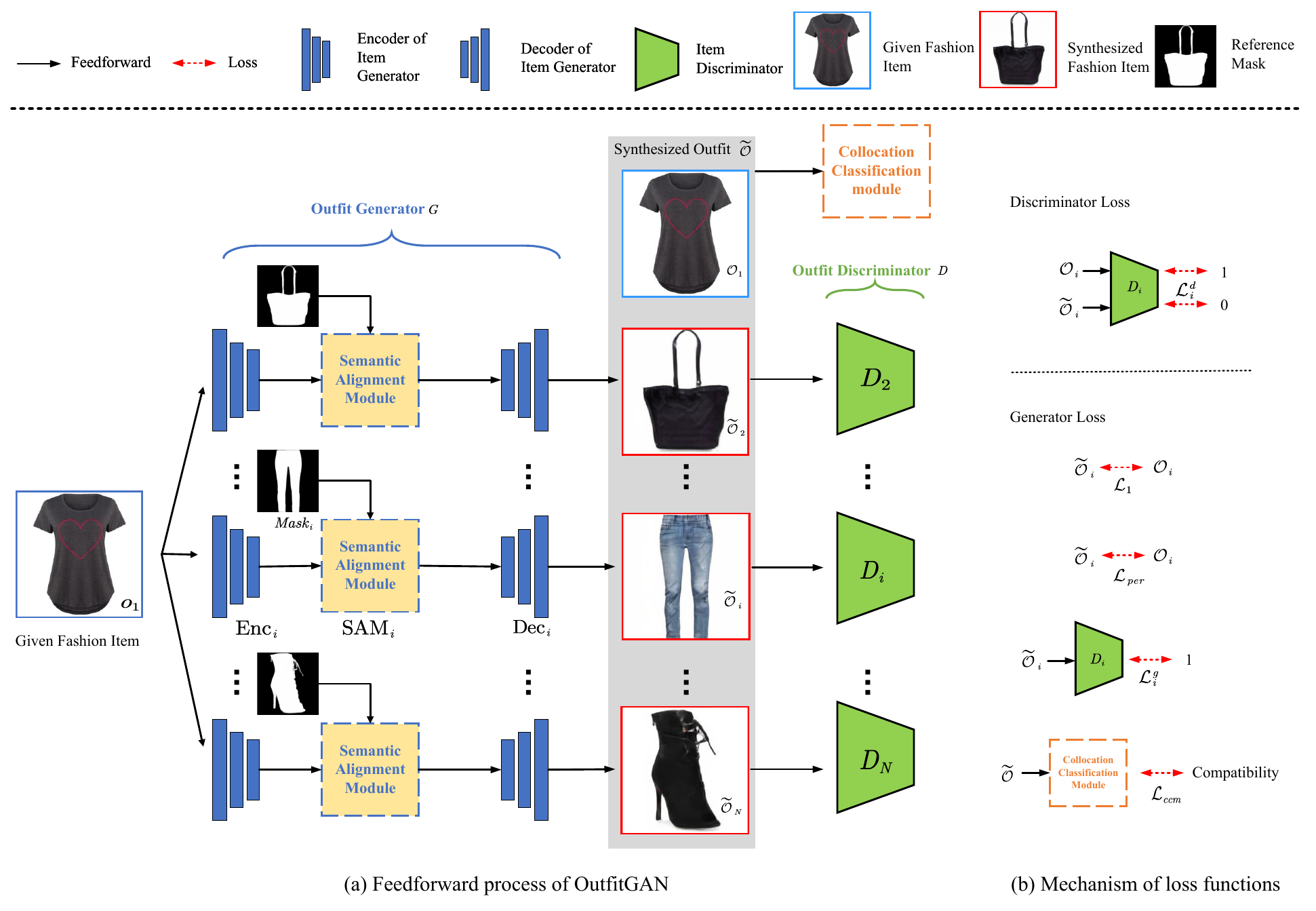}
    \caption{The structure of OutfitGAN framework, which sets $\mathcal{O} _1(k=1)$  as the given fashion item, including: (a) the feedforward process of our OutfitGAN during training and (b) training losses of our model.}
    \label{outfitgan}

\end{figure*}
Based on the problem formulation presented above, we design a new generative framework called OutfitGAN to accomplish the task of outfit generation. The detailed structure of OutfitGAN is illustrated in Fig. \ref{outfitgan}. In particular, Fig. \ref{outfitgan}(a) shows three key modules: an outfit generator $G$, an outfit discriminator $D$, and a collocation classification module CCM. For clarity, the outfit generator and collocation classifier that make up the key components of our OutfitGAN are described firstly. The training losses of our model are shown in Fig. \ref{outfitgan}(b), and are elaborated later in this section. 

\subsubsection{Outfit Generator}
To synthesize a set of complementary items to make up an outfit, our framework needs to learn a mapping function from extant fashion items to new synthesized ones, by considering the compatibility between fashion items. To accomplish this, we train an outfit generator $G$ to translate a given fashion item into multiple collocated ones. In particular, to synthesize a whole outfit that includes  $N$ fashion items, we introduce an outfit generator $G$, which includes  ($N-1$) item generators to synthesize a set of fashion items, conditioned on a given item. The $i$-th item generator $G_i$ includes three components, as shown in Fig. \ref{outfitgan}(a): an encoder $\mathrm{Enc}_i$, a decoder $\mathrm{Dec}_i$, and a semantic alignment module $\mathrm{SAM}_i$. Here, we employ  $\mathrm{Enc}_i$ and $\mathrm{Dec}_i$ in a similar way to a general image-to-image translation generator \cite{wang2018pix2pixHD}. The detailed structures of $\mathrm{Enc}_i$ and $\mathrm{Dec}_i$ can be found in \cite{wang2018pix2pixHD}. The $\mathrm{SAM}_i$ was developed to capture the correspondence between the input and output images. In a compatible outfit, harmonizing elements or styles are often shared by each fashion item. To characterize these shared elements or styles, the $\mathrm{SAM}_i$ is used to capture the collocation correspondences among the fashion items in a certain outfit.
\begin{figure}[t]
    \centering
    \setlength{\abovecaptionskip}{0.cm}
    \includegraphics[width=0.5\textwidth]{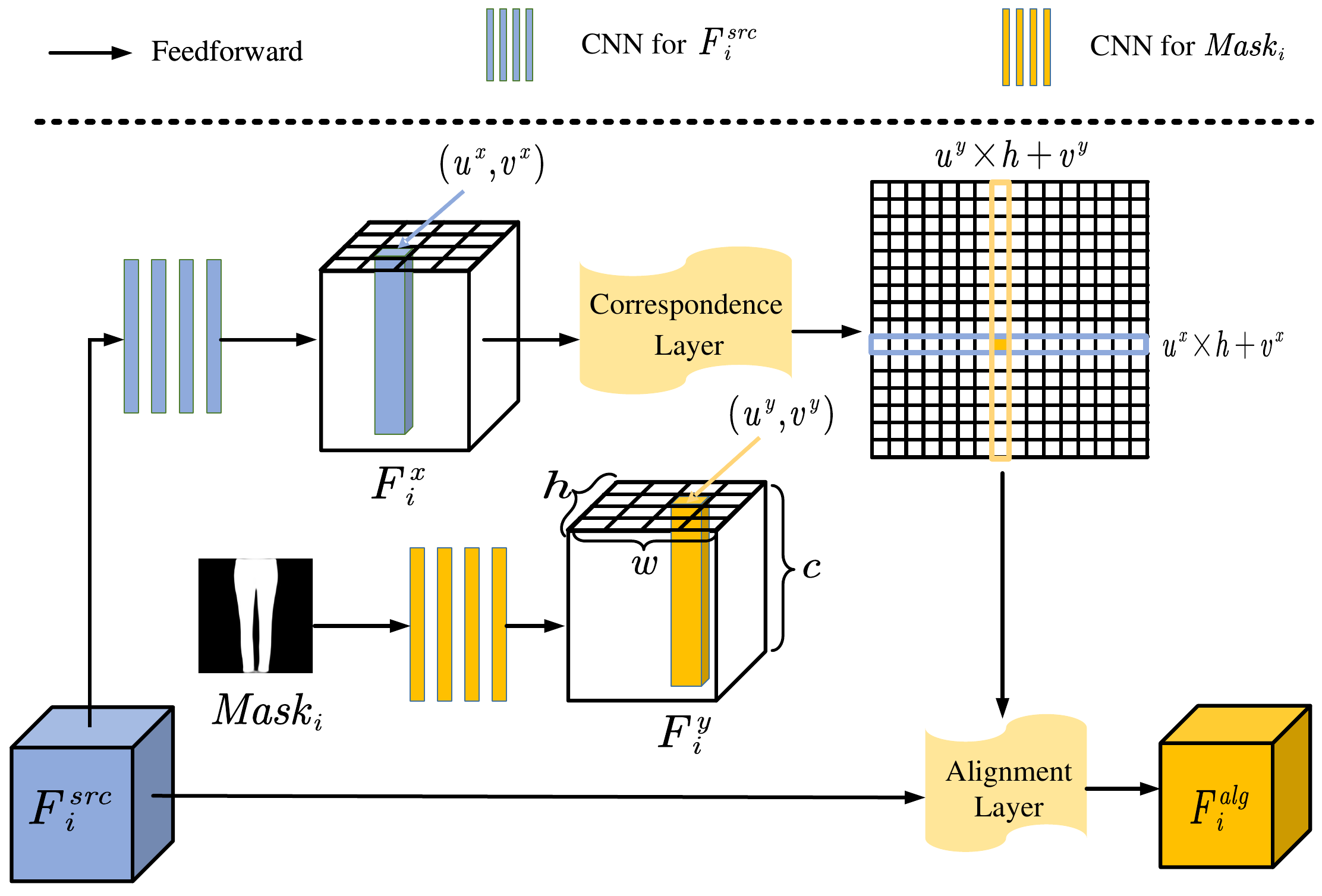}
    \caption{Feedforward process of semantic alignment module.}
    \label{sam}

\end{figure}
In order to fully capture the spatial mapping relationships between the given fashion items and synthesized ones, we use $\mathrm{SAM}_i$ to learn these relationships during the training of OutfitGAN. As shown in Fig. \ref{sam}, $\mathrm{SAM}_i$ has four components: two branches consisting of CNNs, a correspondence layer and an alignment layer. These two CNN branches are used for feature extraction, while the correspondence layer with a differentiable module \cite{rocco2017convolutional} is used to calculate the degree of spatial correlation for each pair of locations for the features extracted by the two CNNs, and the alignment layer aligns the features from $\mathrm{Enc}_i$ based on the degree of spatial correlation. 
We first use two separate CNNs to extract the features of a given fashion item and the reference mask of a target item, which are denoted by $F_{i}^{x}$ and $F_{i}^{y}$, respectively.
 Here, $F_{i}^{x}$ and $F_{i}^{y}$ all lie in the space $\mathbb{R} ^{h\times w \times c}$, where $h$ and $w$ are the height and width of $F_{i}^{x}$ and $F_{i}^{y}$, respectively, and  $c$ is the number of channels. The correspondence layer is then applied to calculate the correspondence matrix $M_{i}^{corr}\in \mathbb{R} ^{(h\times w)\times (h\times w)}$  for these two types of features. In particular, the operation used by the correspondence layer to obtain the correspondence matrix $M_{i}^{corr}$ can be expressed as follows:
\begin{equation}
	M_{i}^{corr}(u,v)=\frac{F_{i}^{x}(u)^TF_{i}^{y}(v)}{\left\| F_{i}^{x}(u) \right\| \cdot \left\| F_{i}^{y}(u) \right\|},
\end{equation}
where $u$ and $v$  are the row and column indexes for $F_{i}^{x}$ and $F_{i}^{y}$, respectively. Each position in $M_{i}^{corr}$ represents the degree of correlation between two positions $F_{i}^{x}$ and $F_{i}^{y}$. As shown in Fig. \ref{sam}, the correlation degree of the  $c$-dimensional feature in $(u^x,v^x)$ of  $F_{i}^{x}$ and that in $(u^y,v^y)$ of  $F_{i}^{y}$ is represented as a scalar value in $(u^x\times h+v^x,u^y\times h+v^y)$ of $M_{i}^{corr}$. A higher value indicates a higher degree of correlation.

After obtaining the correspondence matrix for the input and output images, the feature of a given fashion item from $\mathrm{Enc}_i$, represented as $F_{i}^{src}$, needs to be aligned according to $M_{i}^{corr}$ in order to synthesize a complementary fashion item. To achieve this, the alignment layer is formulated as follows:
\begin{equation}
	F_{i}^{alg}(u)=\sum_v{F_{i}^{src}}(u)\cdot \mathop {softmax} \limits_{v}(M_{i}^{corr}(u,v)),
\end{equation}
where $F_{i}^{alg}$ is the aligned feature from $F_{i}^{src}$ based on $M_{i}^{corr}$. The vector at each position in $F_{i}^{alg}$ is the result of a weighted summation of the vectors in the source feature $F_{i}^{src}$. We then feed $F_{i}^{alg}$ into the decoder $\mathrm{Dec}_i$ to synthesize a fashion item $\widetilde{\mathcal{O} }_i$. 

\subsubsection{Collocation Classification Module}
In this subsection, we describe the CCM, which is used to model the compatibility prediction in order to supervise the compatibility during the outfit generation process. 

More specifically, to ensure that the synthesized outfits fall into the collocation domain, we pre-train a CCM (see Section \ref{ad_train}), which is leveraged to identify whether or not a synthesized outfit is compatible. During the training of OutfitGAN, we fix the parameters of the pre-trained CCM to supervise the compatibility of synthesized items. If the synthesized items are compatible, the CCM applies a smaller penalty to the outfit generator $G$, and if not, the penalty is larger. In particular, the CCM is designed as a sequence model that regards the outfit as a sequence from the perspective of human vision \cite{han2017learning}. To synthesize compatible outfits, we employ a pre-trained sequence model to maintain the compatibility for outfit generation. This includes a pre-trained CNN and two directional LSTMs \cite{bi_lstm}, in order to supervise the compatibility from two directions.
\begin{figure}[t]
    \centering
    \setlength{\abovecaptionskip}{0.cm}
    \includegraphics[width=0.5\textwidth]{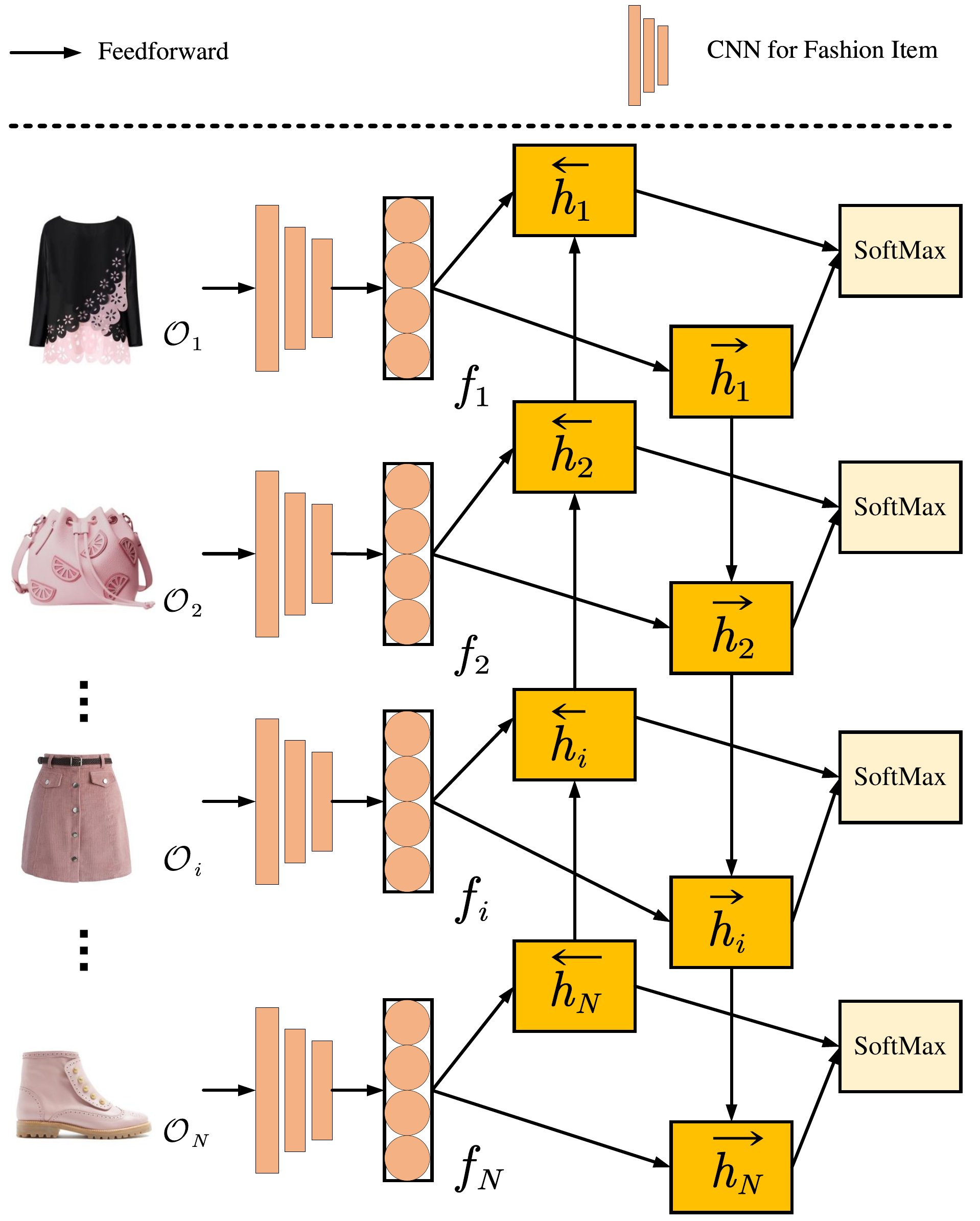}
    \caption{Illustration of collocation classification module.}
    \label{CCM}

\end{figure}
Formally, given a fashion outfit $\mathcal{O} =[\mathcal{O} _1,\cdots ,\mathcal{O} _i,\cdots ,\mathcal{O} _N]$, we regard it as a sequence, where  $\mathcal{O} _i$ is the $i$-th fashion item in $\mathcal{O}$. As shown in Fig. \ref{CCM}, we first extract the latent feature  $f_i$ for $\mathcal{O} _i$  using a pre-trained CNN, and this is then fed into a Bi-LSTM module. For example, the forward LSTM recurrently takes the feature $f_{i-1}$ and the last hidden state $\overrightarrow{h_{i-1}}$ as input and outputs a hidden state  $\overrightarrow{h_i}
$ from  $i=2$ to $N$, as follows:
\begin{equation}
	\overrightarrow{h_i}=LSTM(f_1,\cdots ,f_{i-1}).
\end{equation}
Similarly, the backward LSTM takes the features in the reverse order and outputs the hidden state $\overleftarrow{h_i}
$ from  $i=N-1$ to $1$. We then attempt to maximize the probability of the next item in the outfit given the previous sequence. More formally, we minimize the following compatibility objective function using a cross-entropy loss \cite{murphy2012machine} in the form:
\begin{equation}
\label{l_ccm}
	\begin{aligned}
	\mathcal{L} _{ccm}=&-\frac{1}{N-1}\sum_{i=2}^N{l}og(\frac{exp(\overrightarrow{h_i}f_i)}{\sum_{f_{\cdot}\in \mathcal{F}}{e}xp(\overrightarrow{h_i}f_{\cdot})})\\
	&-\frac{1}{N-1}\sum_{i=N-1}^1{l}og(\frac{exp(\overleftarrow{h_i}f_i)}{\sum_{f_{\cdot}\in \mathcal{F}}{e}xp(\overleftarrow{h_i}f_{\cdot})}),\\
\end{aligned}
\end{equation}
where these two loss terms represent the probabilities of the predictions from the forward and backward LSTM, respectively, and $\mathcal{F}$  denotes all the features $f_{\cdot}$ of the current batch. 
In the pre-training phase, all of the parameters involved in the CCM are learnable. We fix all the learned parameters of the pre-trained module during the training of OutfitGAN. 

\subsubsection{Training Losses}
In addition to the components of OutfitGAN mentioned above, the training losses are of the utmost importance in terms of supervising the training process. As shown in Fig. \ref{outfitgan}(b), the loss function for OutfitGAN includes two types of losses: the outfit discriminator loss and the outfit generator loss. We first discuss the adversarial training loss for the outfit discriminator. As shown in Fig. \ref{outfitgan}(a), our outfit discriminator uses  ($N-1$) independent item discriminators to guide the outfit generation. In a similar way to MUNIT \cite{huang2018munit}, for each item discriminator  $D_i$ we adopt a multi-scale discriminator architecture \cite{wang2018pix2pixHD} and an LSGAN objective \cite{lsgan} to guide the training of our generator. We take the discriminator for the  $i$-th item $\mathcal{O} _i$  as an example. We first downsample the real and synthesized images by factors of two and four. The item discriminator $D_i=\{D_{i,1},D_{i,2},D_{i,3}\}$ is then applied to distinguish between the real and synthesized images at three different scales. Formally, the objective function of our adversarial loss for the training of each item discriminator is expressed as follows:
\begin{equation}
\label{l_i_d}
	\mathcal{L} _{i}^{d}=\sum_{s=1}^3{\mathcal{L} _{i,s}^{d}}(D_i),
\end{equation}
where  $D_i=\{D_{i,1},D_{i,2},D_{i,3}\}$ is the discriminator of the real fashion item  $\mathcal{O} _i$ and the synthesized fashion item  $\widetilde{\mathcal{O} }_i$, and  $\mathcal{L} _{i,s}^{d}$ is the LSGAN objective for training  $D_i$ at a down-sampling scale  $s$. In particular, for each scale  $s$, the $\mathcal{L} _{i,s}^{d}$ is formulated as follows:
\begin{equation}
	\begin{aligned}
	\mathcal{L} _{i,s}^{d}&(D_{i,s})\\
	=&\mathbb{E} _{\mathcal{O} _i\sim p_{data}(\mathcal{O} _i)}(D_{i,s}(\bm{\bigtriangledown} (\mathcal{O} _i,s-1)-1)^2+\\
	&\mathbb{E} _{\mathcal{O} _k\sim p_{data}(\mathcal{O} _k)}(D_{i,s}(\bm{\bigtriangledown} (G_i(\mathcal{O} _k,Mask_i),s-1))^2,\\
\end{aligned}
\end{equation}
where $p_{data}$ is the distribution of real data, $\bm{\bigtriangledown} (x,s)$  represents down-sampling an image $x$  by a factor of $2^s$ , and  $\mathcal{O} _k$ is the given fashion item.

In addition to the outfit discriminator loss, as shown in Fig. \ref{outfitgan}(b), the loss used in the outfit generator has four parts: the adversarial loss for the generator ($\mathcal{L} ^g$) \cite{NIPS2014_gan}, the L1 loss ($\mathcal{L} _1$) \cite{pix2pix2017}, the perceptual loss ($\mathcal{L} _{per}$) \cite{ledig2017photo}, and the CCM loss ($\mathcal{L} _{ccm}$). More specifically, the objective function for our outfit generator loss is defined as follows:
\begin{equation}
\label{l_total}
	\mathcal{L} _{total}=\frac{1}{N-1}\sum_{\substack{i=1 \\i\ne k}}^N{\mathcal{L} _{i}^{g}}+\lambda _1\mathcal{L} _1+\lambda _2\mathcal{L} _{per}+\mathcal{L} _{ccm},
\end{equation}
where $\lambda _1$ and $\lambda _2$  are two coefficients used to balance each loss. In the following, we introduce these losses used in the outfit generator. The compatibility loss was introduced in Eq. (\ref{l_ccm}) and is not discussed here.

\textbf{Adversarial loss}: The objective function of the adversarial loss used in the training of  $G$ includes  ($N-1$) losses for each $G_i$. Each adversarial loss for $G_i$ can be expressed as follows:
\begin{equation}
\label{l_i_g}
	\begin{aligned}
	\mathcal{L} _{i}^{g}&(G_i)=\\
	&\sum_{s=1}^3{\mathbb{E} _{\mathcal{O} _k\sim p_{data}(\mathcal{O} _k)}}[D_{i,s}(\bm{\bigtriangledown} (G_i(\mathcal{O} _k,Mask_i),s-1))-1]^2,\\
\end{aligned}
\end{equation}
where  $\mathcal{L} _{i}^{g}$ is the LSGAN objective function for $G_i$  and  $p_{data}$ is the distribution of real data.

\textbf{L1 loss}: To minimize the difference between the target outfits and the synthesized ones, we use a reconstruction loss (L1) to capture the overall structure of the images from the target domain. Specifically, we keep the discriminator unchanged and add the L1 loss to calculate the absolute distance between the synthesized images and the target ones \cite{pix2pix2017}. This is defined as follows:
\begin{equation}
\label{l_1}
	\mathcal{L} _1=\frac{1}{N-1}\sum_{\substack{i=1 \\ i\ne k}}^N{|}|\widetilde{\mathcal{O} }_i-\mathcal{O} _i||_1,
\end{equation}
where  $\widetilde{\mathcal{O} }_i\in \mathbb{R} ^{256\times 256\times 3}$ denotes a synthesized image of the  $i$-th fashion item and $\mathcal{O} _i\in \mathbb{R} ^{256\times 256\times 3}$ is a target image for the same category.

\textbf{Perceptual loss}: Unlike the L1 loss, the perceptual loss \cite{ledig2017photo} is introduced to ensure that the synthesized images are close to the target ones in high-level feature space. It also measures the perceptual difference between the images in terms of their content and style. 
Here, we adopt the perceptual loss to ensure that our OutfitGAN produce images that are similar to the ground truths. We compute the perceptual loss in the 
\textit{relu1\_2}, \textit{relu2\_2}, \textit{relu3\_3} and \textit{relu4\_3} layers of the VGG-16 
network $\phi $  which was pre-trained on ImageNet \cite{image_net}. We then apply an auxiliary benchmark from DeepFashion \cite{liu2016deepfashion} consisting of 50 categories of fashion items with 289,229 images, each of which is annotated with 1,000 descriptive attributes. This benchmark was used to classify the attributes, in order to fine-tune our network $\phi$. Specifically, the perceptual loss adopted here is defined as:
\begin{equation}
\label{l_per}
	\mathcal{L} _{per}=\frac{1}{N-1}\sum_{\substack{i=1\\i\ne k}}^N{\sum_l{|}}|\phi _l(\widetilde{\mathcal{O} }_i)-\phi _l(\mathcal{O} _i)||_1,
\end{equation}
where $l\in \{relu1\_2,relu2\_2,relu3\_3,relu4\_3\}$ is the aforementioned layer of VGG-16, and  $\phi _l$ represents the function of layer $l$.
\subsection{Implementation Details}
In this subsection, we introduce two reference mask generation strategies used to synthesize the alternative masks. Pix2Pix mask generation is used to synthesize reference masks via a pre-trained generative model, and random mask generation is used to return the sampled reference masks from the training set. In this following, we discuss the details of the detailed network architecture of OutfitGAN. Finally, we illustrate the overall adversarial algorithm used to train OutfitGAN.
\subsubsection{Strategies for Reference Mask Generation}
\label{mask_gen}
The reference mask is an essential component of OutfitGAN, and provides important guidance information in terms of supervising the generation of compatible fashion items. In addition to the reference masks given by users, we may in practice need to synthesize reference masks based on models, such that they can then be fed into OutfitGAN. To overcome this issue, we design two strategies for the synthesis of a diverse range of masks: Pix2Pix and random mask generation. These two methods mimic the phase in which fashion designers or common users generate reference masks, and extend the basic functions of OutfitGAN. 

\textbf{Mask generation using Pix2Pix}: In order to take advantage of reference masks to improve the effectiveness of OutfitGAN, we propose a method of synthesizing reference masks using a pre-trained generative model to extend the function of OutfitGAN. In fact, reference mask generation can be regarded as another image-to-image translation task, in which the input is an RGB image of a given fashion item and the output consists of the corresponding masks of compatible fashion items. There are many methods that are capable of synthesizing reference masks for a given fashion item \cite{pix2pix2017}\cite{wang2018pix2pixHD}. In this research, Pix2Pix \cite{pix2pix2017}, as a representative framework among these methods, was chosen for this task. We constructed a large-scale dataset called OutfitSet (more details are given in Section \ref{dataset_construct}), which consisted of 20,000 outfits with their associated masks, and used this to train the mask generator. Using ($N-1$) fashion items for each complete outfit (i.e., excluding the given fashion item), we pre-trained ($N-1$) independent Pix2Pix mask generators on the training set of OutfitSet to synthesize the reference masks for the target fashion items. For example, given an RGB upper clothing image, we needed to synthesize the corresponding masks for the bag, lower clothing and shoes. Each mask generator included an encoder, several residual blocks and a decoder. Each discriminator used the PatchGAN architecture \cite{pix2pix2017}, and the LSGAN \cite{lsgan} and L1 losses were employed to guide the process of reference mask generation. We used the default setting given in \cite{pix2pix2017} to pre-train the Pix2Pix mask generator. The detailed structure of the network can be found in \cite{pix2pix2017}. In the testing phase of OutfitGAN, the reference masks for compatible fashion items can be automatically synthesized by the pre-trained Pix2Pix mask generator without the need for assistance from users. 

\textbf{Random mask generation}: In a real-world clothing collocation application, users can use different reference masks to guide the generation of outfits. In order to meet these personalized requirements, we can use a random mask generator that randomly selects a reference mask for a compatible fashion item from a source dataset. In our implementation, the training set for our constructed OutfitSet was used as a source dataset for reference masks. Given a fashion item, reference masks corresponding to target compatible fashion items can be randomly selected from the source dataset based on the categories of target collocation items. This strategy reflects the different tastes of users in terms of the selection of reference masks, and increases the diversity of the generated compatible fashion items to some extent. In our experiments, we also performed a detailed empirical study of the effects of different reference mask generation strategies on the results.

\subsubsection{Network Architecture}
\begin{figure}[t]
    \centering
    \setlength{\abovecaptionskip}{0.cm}
    \includegraphics[width=0.51\textwidth]{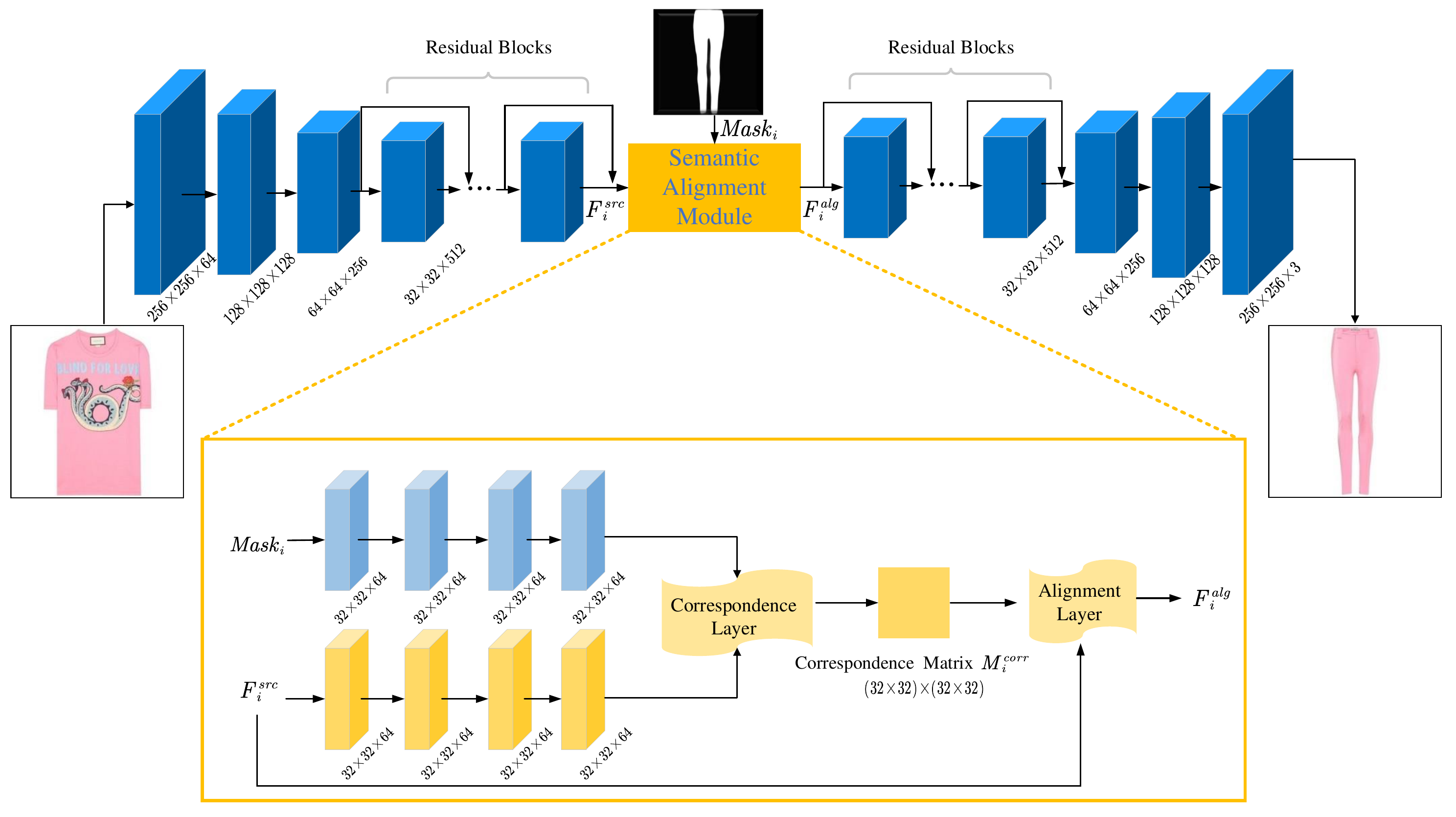}
    \caption{Network architecture of the $i$-th item generator $G_i$.}
    \label{network_det}

\end{figure}
In this subsection, we describe the detailed network architecture of OutfitGAN. For the  $i$-th item generator $G_i$, as shown as an example in Fig. \ref{network_det}, we employ the architecture of encoder $\mathrm{Enc}_i$ and the decoder $\mathrm{Dec}_i$ from \cite{huang2018munit}, in which their effectiveness in image-to-image translation is proven \cite{CycleGAN2017}. The encoder $Enc_i$ includes four convolutional blocks (conv-blocks) and three residual blocks (res-blocks), whereas the decoder contains three res-blocks, three upsampling and conv-block modules, and one conv-block followed by a Tanh function. We apply a ReLU activation function to all the conv-blocks. As illustrated in Fig. \ref{network_det}, the SAM applies four conv-blocks to each branch of the feature extractor, followed by a correspondence layer and an alignment layer. Our discriminator is designed using a multi-scale architecture, in the same way as in \cite{huang2018munit}. In the CCM, we extract image features with a pre-trained ResNet-50 \cite{2015arXiv151203385H} provided by PyTorch \cite{paszke2017automatic} and a fully-connected network for 512-dimensional embeddings. A Bi-LSTM (which includes forward and backward LSTM) is then used to model the collocation relationship. In the same way as in \cite{wang2019diagnosis}, the number of layers for each LSTM is set to one, and the number of hidden features is set to 512.
\subsubsection{Adversarial Training Process}
\label{ad_train}
\begin{algorithm}[t]
\small
  \caption{Adversarial training algorithm for OutfitGAN}
  \label{algorithm}
  \SetAlgoLined
  \KwIn{Extant fashion item $\mathcal{O}_{k}$, reference masks $[Mask_{1}, \cdots,Mask_{k-1},Mask_{k+1}\cdots, Mask_{N}]$, and target fashion items $[\mathcal{O}_{1},\cdots,\mathcal{O}_{k-1},\mathcal{O}_{k+1},\cdots,\mathcal{O}_{N}]$}
  \KwOut{OutfitGAN generator $G$}
  Pre-train collocation classification module CCM on our training set and select the best model through validation set, update the parameters $\theta_{\mathrm{CCM}}$ of CCM with
  
  \quad$\theta_{\mathrm{CCM}}\leftarrow\theta_{\mathrm{CCM}}-\eta_{\mathrm{CCM}}\bigtriangledown_{\theta_{\mathrm{CCM}}}(\mathcal{L} _{ccm})$; // See {Eq. (\ref{l_ccm})}\\
  Fine-tune the VGG-16 for attributes classification on DeepFashion; // Prepare for perceptual loss\\
  Initialize the parameters $\theta_{G}$, $\theta_{D}$ of $G$, $D$, respectively; fix all parameters of CCM and VGG-16; \\
  
  \For{$iter\leftarrow 1$ \KwTo $N_{iter}$}{
  sample a batch of $\mathcal{O}=[\mathcal{O}_{1},\cdots,\mathcal{O}_{N}]$ and reference masks $\{Mask_{1},\cdots,Mask_{N}\}$ from training set\;
  \For{ $i\in\{1,\cdots,k-1,k+1,\cdots,N\}$}{
  $\mathcal{L}^{d}_{i}\leftarrow \sum_{s=1}^3{\mathcal{L} _{i,s}^{d}}(D_i)$; \quad // See  {Eq. (\ref{l_i_d})}
  
  update $\theta_{D_i}\in \theta_D$ with
  
  \quad$\theta_{D_i}\leftarrow\theta_{D_i}-\eta\bigtriangledown_{\theta_{D_i}} (\mathcal{L}^{d}_{i})$;} 
  
  \For{ $i\in\{1,\cdots,k-1,k+1,\cdots,N\}$}{
  $\mathcal{L}^{g}_{i}\leftarrow \sum_{s=1}^3{\mathcal{L} _{i,s}^{g}}(D_i)$; \quad // See  {Eq. (\ref{l_i_g})}
  }
  $\mathcal{L} _1\leftarrow\frac{1}{N-1}\sum_{\substack{i=1 \\i\neq k}}^N{|}|\widetilde{\mathcal{O}}_i-\mathcal{O}_i||_1$; \quad // See  {Eq. (\ref{l_1})}

$\mathcal{L} _{per}=\frac{1}{N-1}\sum_{\substack{i=1 \\i\neq k}}^N{\sum_l{|}}|\phi _l(\widetilde{\mathcal{O}}_i)-\phi _l(\mathcal{O}_i)||_1$;\quad // See  {Eq. (\ref{l_per})}

$\mathcal{L} _{ccm}\leftarrow-\frac{1}{N-1}\sum_{i=2}^N{l}og(\frac{exp(\overrightarrow{h_i}f_i)}{\sum_{f_{\cdot}\in \mathcal{F}}{e}xp(\overrightarrow{h_i}f_{\cdot})})
	-\frac{1}{N-1}\sum_{i=N-1}^1{l}og(\frac{exp(\overleftarrow{h_i}f_i)}{\sum_{f_{\cdot}\in \mathcal{F}}{e}xp(\overleftarrow{h_i}f_{\cdot})})$;  // See  {Eq. (\ref{l_ccm})}

  update $\theta_{G}$ with
  
  \quad$\theta_{G}\leftarrow\theta_{G}-\eta\bigtriangledown_{\theta_{G}}(
\frac{1}{N-1}\sum^{N}_{\substack{i=1 \\i\neq k}}{\mathcal{L} _{i}^{g}}+\lambda _1\mathcal{L} _1+\lambda _2\mathcal{L} _{per}+\mathcal{L} _{ccm}
)$; \qquad // See  Eq. (\ref{l_total})
}

\end{algorithm}

In this subsection, we present the design of an adversarial training scheme which is used to optimize the generator $G$ of OutfitGAN. For clarity, the entire training process of OutfitGAN is summarized in Algorithm \ref{algorithm}. We first pre-train the collocation classification module on our training set by minimizing the loss in Eq. (\ref{l_ccm}) with a learning rate  $\eta _{ccm}$. We then select the best CCM model with our validation set (see Section \ref{dataset_construct}) by calculating the smallest $\mathcal{L} _{ccm}$  according to Eq. (\ref{l_ccm}) (shown in lines 1-2). Following this, we fine-tune the VGG-16, which was pre-trained on ImageNet, by applying the attribute classification method used in DeepFashion (shown in line 3). We initialize the parameters of $G$ and $D$, and fix all the parameters of the CCM and VGG-16 during the training of OutfitGAN (shown in line 4). The subsequent training process is carried out by applying a gradient descent step to $D$ and $G$ in alternate steps, and using the gradient descent method to update the parameters  $\theta _D$ and $\theta _G$ of $D$ and $G$, respectively (shown in lines 5-20). Specifically, given a batch of outfit images and reference masks, we train each $D_i\in D$  by reducing the loss in Eq. (\ref{l_i_d}) (shown in line 10). We then fix $\theta _D$  and calculate the adversarial loss ($\mathcal{L} _{i}^{g}$) for each  $G_i\in G$ (shown in line 13) and L1 loss ($\mathcal{L} _1$) (shown in line 15), the perceptual loss ($\mathcal{L} _{per}$) (shown in line 16) and the CCM loss ($\mathcal{L} _{ccm}$) (shown in line 17) for $G$. Finally, we optimize $\theta _G$  by reducing the loss in Eq. (\ref{l_total}) (shown in line 19). We train $D$ and  $G$ over $N_{iter}$ iterations with a learning rate of $\eta$.
\section{Experiments}
\label{exp}
In this section, we first describe the construction of our dataset in detail. Parameter settings of models and evaluation metrics are then described sequentially. The performance of our proposed OutfitGAN is compared against several competitive image-to-image translation baselines, and we perform an ablation study to verify the effectiveness of the main modules in OutfitGAN. Furthermore, we conduct a parametric study on our model and an extra study on different sequences of fashion items used in the collocation classification module. Finally, the limitation of our framework is discussed.

\subsection{Dataset}
\label{dataset_construct}
\begin{figure}[t]
    \centering
    \setlength{\abovecaptionskip}{0.0cm}
    \includegraphics[width=0.5\textwidth]{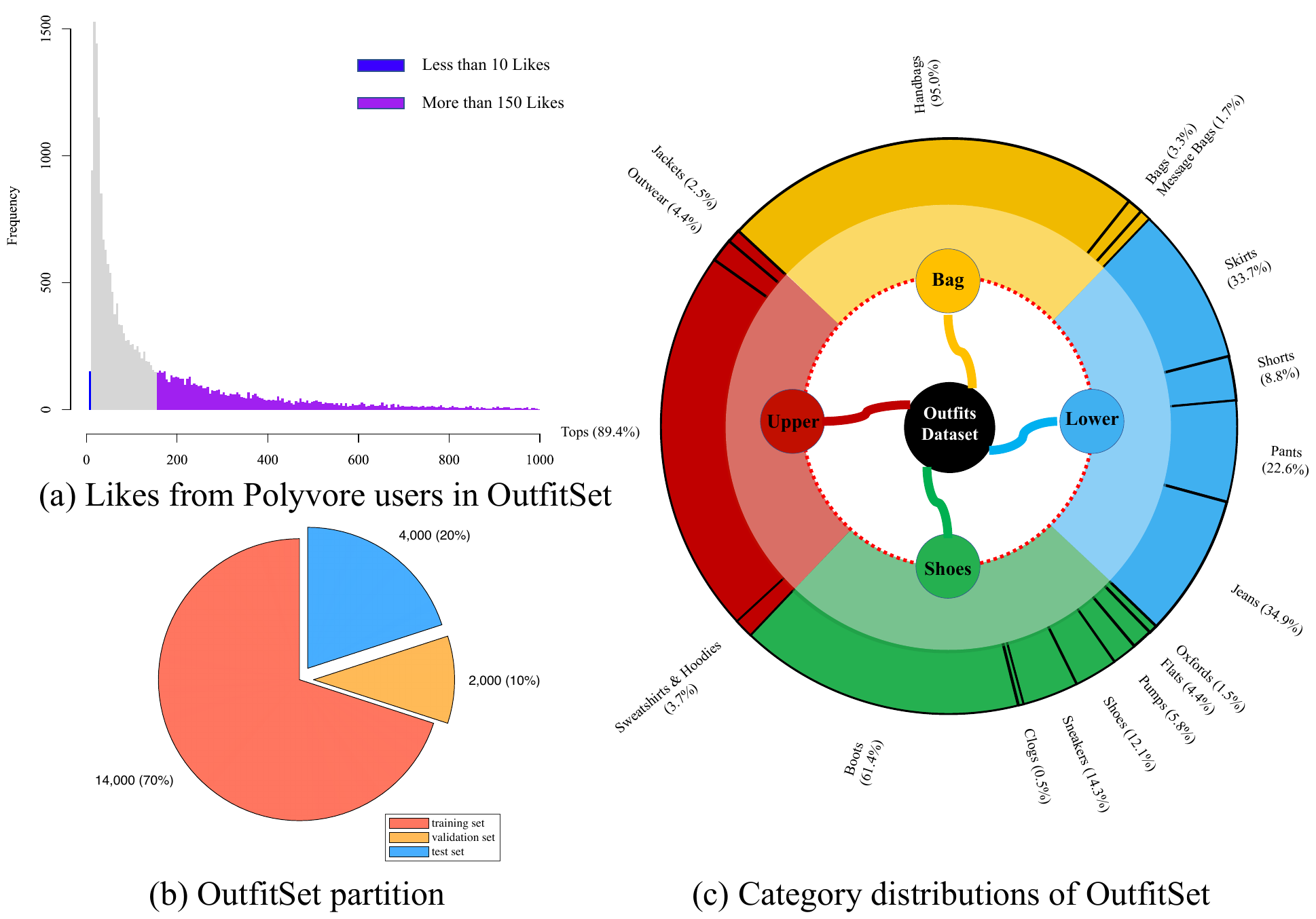}
    \caption{Statistics of OutfitSet.}
    \label{dataset_stat}

\end{figure}

When carrying out fashion outfit generation, accurate fashion datasets are of the utmost importance in terms of providing the ground truths for model training and evaluation. Although many public fashion outfit compatibility datasets are available for fashion modeling, such as UT Zappos50K \cite{yu2014fine}, the Maryland Polyvore dataset \cite{han2017learning}, FashionVC \cite{fashionvc}, and IQON3000 \cite{10.1145/3343031.3350956}, all of these lack explicit common category annotations for fashion items and clear compositions for outfits. To overcome these issues in current datasets and to verify the effectiveness of our proposed outfit generation models, we collected fashion outfits from a fashion matching website, Polyvore.com, which contained numerous compatible outfits constructed by fashion experts. These outfits were put together based on the preferences of fashion experts, with the aim of clearly and attractively presenting specific fashion styles. The original dataset consisted of over 344,560 outfits, which were composed of 2,131,607 fashion items. We selected four types of fashion items (upper clothing, bag, lower clothing and shoes) that are common components of outfits worn in daily life. We used the upper clothing as the given fashion item in order to exploit the richer information on styles that can be obtained from the upper clothing compared with the other fashion items in the same outfit. This means that for each outfit set   [upper clothing, bag, lower clothing, shoes] (i.e., $N=4$), the extant given fashion item   represents upper clothing. We therefore kept only those outfits that included all four of these categories. Since images of shoes have diverse orientations due to the different shooting angles used, we filtered out images that only contained one shoe, and flipped all left-oriented shoes horizontally to form right-oriented shoes. After this process was complete, 32,043 outfits remained in the dataset. We selected the top 20,000 outfits based on the number of likes given by Polyvore users. As shown in Fig. \ref{dataset_stat}(a), most of the outfits had more than 10 likes and less than 150. We then partitioned these outfits randomly into three folds, to form a training set, a validation set and a test set, and these constituted our OutfitSet dataset. The training set contained 14,000 outfits (70\%), the validation set 2,000 outfits (10\%) and the test set 4,000 outfits (20\%), as shown in Fig. \ref{dataset_stat}(b). In particular, the training set was used to train the CCM, OutfitGAN, and the Pix2Pix mask generation strategy; the validation set was used to select the best pre-trained CCM; and the test set was used to evaluate OutfitGAN at the testing stage. It is worth noting that the OutfitSet dataset also contained many sub-classes for each class, as illustrated in Fig. \ref{dataset_stat}(c), and had relatively rich category-based annotations in comparison to existing datasets. For automatic reference mask generation (see Section \ref{mask_gen}), we employed a saliency detector \cite{Liu_2019_CVPR} to detect the masks of fashion items for use as reference masks to guide the mask generation. Each fashion item in our dataset had a corresponding mask in the form  $[m]^{256\times 256},m\in \{0,1\}$, where a value of one denotes the segmentation of fashion items and zero denotes the background of an image.
\subsection{Experimental Setup and Parameter Settings}
In the experiments, all images were resized to $256\times256$, and we used random cropping for data augmentation during training. In the training phase, the batch size was set to four, and the number of training iterations for the model was set to 200,000. All experiments were performed on a single NVIDIA GeForce RTX 3090, and the implementation was carried out in PyTorch \cite{paszke2017automatic}. We set the coefficients to balance the losses as follows:  $\lambda _1=100$ and  $\lambda _1=10$ for OutfitGAN with real reference masks, and  $\lambda _1=10$ and  $\lambda _1=10$ for OutfitGAN with reference masks produced by mask generation strategies. The CCM was trained with an SGD \cite{zeiler2012adadelta} optimizer with a learning rate $\eta_{cmp}$ of 0.2 and a momentum of 0.9. OutfitGAN was trained with an Adam \cite{kingma2014adam} optimizer with $\beta _1=0$  and $\beta _2=0.99$, and the learning rate  $\eta$ for $G$ and $D$ was set to $10^{-4}$.
\subsection{Evaluation Metrics}
To evaluate the performance of our proposed model, we used a variety of evaluation metrics from three perspectives, as follows:
\begin{enumerate}
\item A similarity measurement was used to measure the similarity between the synthesized images and the target ones. We adopted two metrics: a structural similarity (SSIM) \cite{wang2004image} and a learned perceptual image patch similarity (LPIPS) \cite{zhang2018perceptual}. SSIM \cite{wang2004image} is a traditional, widely used image quality index for image comparison. Given two local patches extracted from input images, i.e., a real image patch  $x$ and a synthesized image patch $y$, SSIM measures the luminance, contrast and similarity of $x$ and $y$, where a higher score indicates a higher similarity. LPIPS \cite{zhang2018perceptual} is another common metric used to evaluate the image similarity between two images, particularly for a synthesized image and a target one, with a pre-trained deep model. We used the default pre-trained AlexNet \cite{alexnet} provided by the authors \cite{zhang2018perceptual} to calculate the LPIPS metric. Here, a higher score indicates a lower similarity, and vice versa. 
\item An authenticity measurement was applied to reflect the quality of the synthesized images in terms of their authenticity. Previous studies \cite{choi2020starganv2} have suggested that the Fréchet inception distance (FID) can be used to estimate the authenticity of synthesized images in feature space. More specifically, the FID measures the similarity between two domains of images, and is particularly suitable for real images and images synthesized by GANs. To calculate the FID between two image domains $\mathcal{Y}$  and $\mathcal{Y}'$, we first embed both images into the same feature space $F$ given by an Inception model \cite{Szegedy_2016_CVPR}. The FID can be defined as follows:
\begin{equation}
\begin{aligned}
	\rm{FID}(\mathcal{Y} ,\mathcal{Y} ')=&||\mu _{\mathcal{Y}}-\mu _{\mathcal{Y} '}||_{2}^{2}+\\
	&Tr(\Sigma _{\mathcal{Y}}+\Sigma _{\mathcal{Y} '}-2(\Sigma _{\mathcal{Y}}\Sigma _{\mathcal{Y} '})^{\frac{1}{2}}),
\end{aligned}
\end{equation}
where $\mu _{\mathcal{Y}}$  and $\mu _{\mathcal{Y}'}$  are the average values of the feature space $F$  for $\mathcal{Y}$ and $\mathcal{Y}'$, respectively; $\Sigma _{\mathcal{Y}}$  and $\Sigma _{\mathcal{Y}'}$  are their variances, respectively; and  $Tr(\cdot)$ is the trace of the matrix. A lower FID score indicates a higher visual authenticity for the synthesized images, and vice versa. 
\item A compatibility measurement was used to gauge the degree of matching between the synthesized outfits. In order to perform a fair evaluation in terms of the compatibility of each outfit, we developed a new metric called the fashion compatibility test score (FCTS). For this metric, we used an open-source toolbox MMFashion\footnote{https://github.com/open-mmlab/mmfashion}  to evaluate the fashion compatibility between the items making up an outfit. The fashion compatibility predictor module of MMFashion was developed on the basis of the work in \cite{vasileva2018learning} on fashion compatibility prediction. To enable a fair comparison, this fashion compatibility predictor $\psi$ was trained on the Maryland Polyvore dataset \cite{han2017learning}, meaning that its training set was different from our OutfitSet, and the pre-trained model was provided by MMFashion. We calculated the FCTS for all models as follows. Firstly, both positive and negative samples were constructed. We defined positive samples as outfits synthesized by generative models, and the negative samples were randomly composed of synthesized fashion items which are not from the same outfit. We assume the synthesized outfits (positive samples) are more compatible than the randomly composed ones (negative samples). We tested the compatibility score between positive and negative samples based on FCTS, which can be defined as:
\begin{equation}
	\rm{FCTS}=\frac{\Sigma _{j=1}^{N_{cmp}}[Comp(outfit_{j}^{p})>Comp(outfit_{j}^{n})]}{N_{cmp}},
\end{equation}
where  $Comp(\cdot)$ is the fashion compatibility score computed by the compatibility predictor $\psi$;  $N_{cmp}$ denotes the number of comparisons between the positive and negative samples; $outfit_{j}^{p}$ and and $outfit_{j}^{n}$  denote a positive and negative outfit sample, respectively. A higher score indicates better compatibility. 
\end{enumerate}
\subsection{Performance Comparison}
\subsubsection{Compared Methods}
To examine the effectiveness of our proposed OutfitGAN, we compared it with six state-of-the-art methods: Pix2Pix \cite{pix2pix2017}, Pix2PixHD \cite{wang2018pix2pixHD}, CycleGAN \cite{wang2018pix2pixHD}, MUNIT \cite{huang2018munit}, DRIT++ \cite{DRIT_plus}, and StarGAN-v2 \cite{choi2020starganv2}. These include both supervised and unsupervised models. For completeness, we give a brief introduction to these methods as follows:

\textbf{Pix2Pix} \cite{pix2pix2017} is the first framework developed for supervised image-to-image translation, and uses a U-Net architecture for the generator and a single discriminator to classify real and fake image pairs.

\textbf{Pix2PixHD} \cite{wang2018pix2pixHD} is an improved version of Pix2Pix framework based on a coarse-to-fine approach, which uses a coarse-to-fine generator, a multi-scale discriminator and a feature matching loss.

\textbf{CycleGAN} \cite{CycleGAN2017} is an unsupervised image-to-image translation method with a cycle reconstruction loss, and was the first framework to address the issue of unpaired image-to-image translation.

\textbf{MUNIT} \cite{huang2018munit} is based on the idea that an image representation can be decomposed into a style code and a content code. It can learn disentangled representations for image-to-image translation.

\textbf{DRIT++} \cite{DRIT_plus} is an improved version of DRIT \cite{DRIT}, which disentangles the latent spaces into a shared content space and an attribute space for each domain and was developed to synthesize diverse images for image-to-image translation.

\textbf{StarGAN-v2} \cite{choi2020starganv2} is an improved version of StarGAN \cite{choi2018stargan} that employs an MLP to synthesize different styles and then injects them into decoders to synthesize a diverse range of images.

It should be noted that except for StarGAN-v2, these baseline methods can synthesize only one target domain image given an image from the source domain. We therefore trained  ($N-1$)  models for each baseline model independently, except for StarGAN-v2. The implementations of these models were all based on original codes released by the authors, and the hyperparameters were tuned to adapt to our OutfitSet.

\subsubsection{Comparison of Results}
\begin{table*}[t]
\centering
\caption{Results of compared methods (here, for all metrics except LPIPS and FID, higher is better)}
\label{ssim_lpips_fid_table}
\begin{tabular}{p{1.84cm} P{1.04cm} P{1.04cm} P{1.04cm} P{0.04cm} P{1.04cm} P{1.04cm} P{1.04cm} P{0.04cm} P{1.04cm} P{1.04cm} P{1.04cm} P{0.04cm} P{1.04cm}}
\toprule
Method & \multicolumn{3}{c}{SSIM($\uparrow$)} & & \multicolumn{3}{c}{LPIPS($\downarrow$)} & & \multicolumn{3}{c}{FID($\downarrow$)}& &{FCTS($\uparrow$)}\\
\cline{2-4} \cline{6-8} \cline{10-12}
                        & bag & lower & shoes & & bag & lower & shoes & & bag & lower & shoes\\
 \hline
 Pix2Pix \cite{pix2pix2017}                      & 0.384 & 0.562 & 0.512 && 0.576 & 0.415 & 0.489&& 58.024&55.399&57.315&&57.5\%\\
 Pix2PixHD \cite{wang2018pix2pixHD} & 0.468 & 0.620 & 0.542 && 0.556&0.365&0.448&&151.683&191.154&134.126&&57.5\%\\
 CycleGAN \cite{CycleGAN2017} & 0.392 & 0.447 & 0.504 && 0.570 & 0.515 & 0.468&& 51.695&56.625&43.497&&{86.4\%}\\
 MUNIT \cite{huang2018munit} & 0.444 & 0.553 & 0.563 && 0.553 & 0.403 &0.443&& 177.605&88.730&85.521&&52.0\%\\
 DRIT++ \cite{DRIT_plus} & 0.392 & 0.470 & 0.514  & & 0.585 & 0.510 & 0.480 & & 74.474 & 108.055 & 80.916 & & 52.2\%\\
 StarGAN-v2 \cite{choi2020starganv2} & 0.355 & 0.610 & 0.477 && 0.590 & 0.355&0.465 && 153.603 & 116.923 & 116.467&&50.3\%\\
 \hline
 OutfitGAN & \textbf{0.643} & \textbf{0.718} & \textbf{0.748} && \textbf{0.385} & \textbf{0.270} & \textbf{0.264} && \textbf{46.183} & \textbf{29.675} &\textbf{36.848}&&\textbf{87.1\%}\\
OutfitGAN-P & --&--&--& &--&--&--&&\textbf{33.541} & \textbf{35.324}&\textbf{36.409}&&\textbf{91.9\%}\\
 OutfitGAN-R & --&--&--& &--&--&--&&\textbf{40.809} & \textbf{41.528}&\textbf{39.152} &&\textbf{91.4\%}\\
\bottomrule
\end{tabular}
\end{table*}

\begin{figure*}[t]
    \centering
    \setlength{\abovecaptionskip}{0.cm}
    \includegraphics[width=1.0\textwidth]{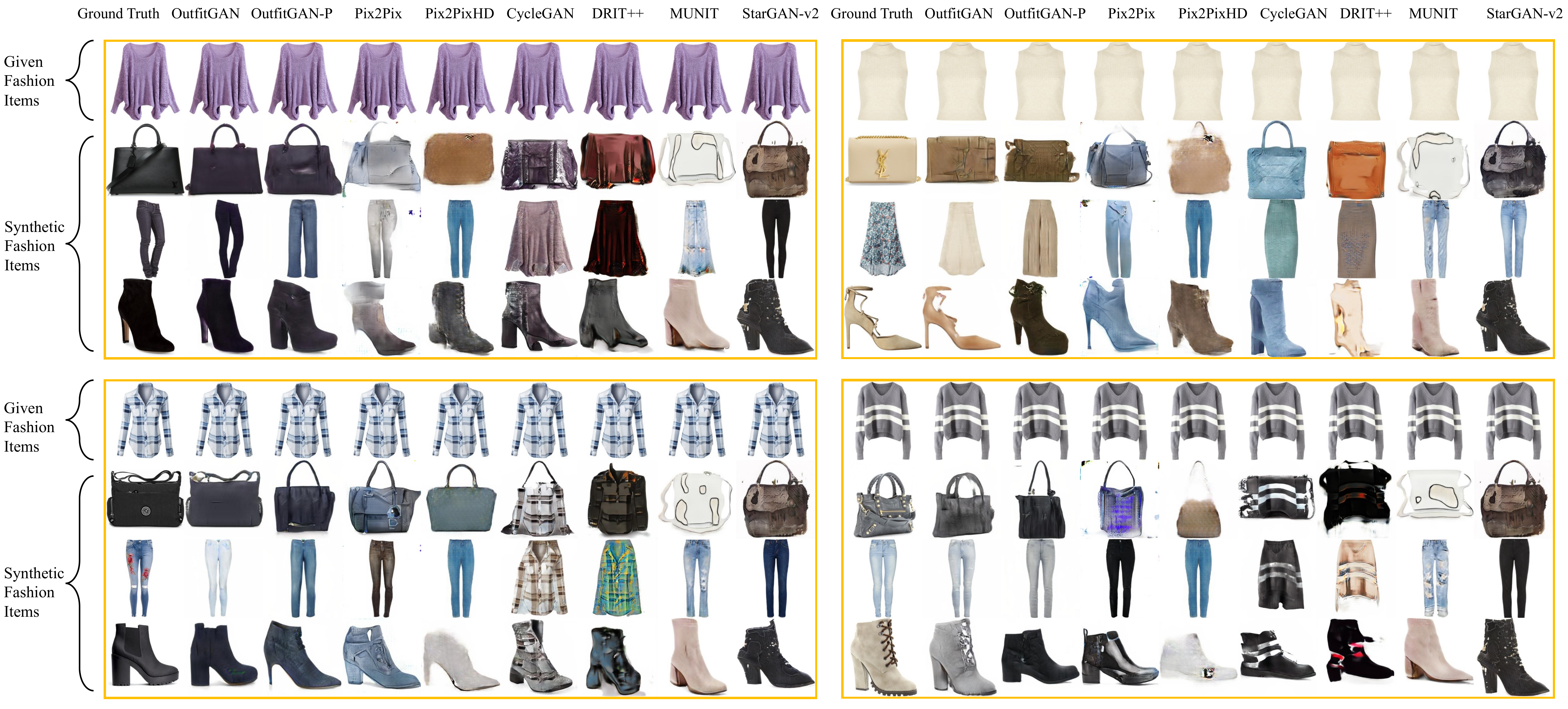}
    \caption{Synthesized samples of our models and baselines.}
    \label{samples}

\end{figure*}

A quantitative comparison of the results for all of the evaluation metrics is given in Table \ref{ssim_lpips_fid_table}. As described in Section \ref{mask_gen}, reference masks that represent the outlines of target fashion items play an important role in guiding our model. These reference masks can be divided into three types based on their source, i.e., whether they were provided by a user, synthesized by a generative model or randomly selected by the system. As shown in Table \ref{ssim_lpips_fid_table}, we use OutfitGAN, OutfitGAN-P, and OutfitGAN-R to denote our model with real reference masks, synthesized reference masks from Pix2Pix mask generation and reference masks from random selection, respectively, in the following discussion. It should be noted that since different reference masks can produce different synthesized fashion items, we compare OutfitGAN-P and OutfitGAN-R with the baselines in terms of only the authenticity and compatibility measures, i.e., FID and FCTS. Table \ref{ssim_lpips_fid_table} shows that our proposed OutfitGAN consistently outperforms other image-to-image translation methods in terms of all three metrics (similarity, authenticity and compatibility). Fig. \ref{samples} shows examples of synthesized fashion items from our models and other baseline models. Since the reference masks of OutfitGAN and OutfitGAN-R are from the same domain, the results of OutfitGAN-R are omitted here. From Fig. \ref{samples}, it is clear that our model produces superior results, particularly in terms of the textural details and the harmony of the styles with the given fashion items. The results of a quantitative evaluation show that OutfitGAN yields approximate performance improvements of 0.175, 0.098, and 0.185 in the SSIM for the categories of bag, lower clothing and shoes, respectively, in comparison to the second-best method; it also reduces approximate 0.168, 0.085, and 0.179 values of LPIPS for the generation of bag, lower clothing, and shoes, respectively, compared with other methods. This suggests that our synthesized images can maintain the overall image structure and visual similarity better than other methods. Fig. \ref{samples} shows that OutfitGAN can synthesize the most similar results in terms of visual appearance. This means that our approach not only surpasses other methods in terms of the quantitative similarity metrics, but also outperforms them in terms of visual observations. 

\begin{table*}[t]
\centering
\caption{Comparative results for OutfitGAN in terms of semantic alignment module}
\label{ssim_lpips_fid_zero_channel_table}
\begin{tabular}{p{2.7cm} P{1.0cm} P{1.0cm} P{1.0cm} P{0.05cm} P{1.0cm} P{1.0cm} P{1.0cm} P{0.05cm} P{1.0cm} P{1.0cm} P{1.0cm}}
\toprule
Method & \multicolumn{3}{c}{SSIM($\uparrow$)} & & \multicolumn{3}{c}{LPIPS($\downarrow$)} & & \multicolumn{3}{c}{FID($\downarrow$)}\\
\cline{2-4} \cline{6-8} \cline{10-12}
                        & bag & lower & shoes & & bag & lower & shoes & & bag & lower & shoes\\
\hline
 OutfitGAN w/o SAM & 0.621 & 0.718 & 0.729 && 0.399 & 0.284 & 0.279 && 48.584 & 34.435 &50.973\\
 \hline
 OutfitGAN w/ SAM & \textbf{0.643} & \textbf{0.718} & \textbf{0.748} && \textbf{0.385} & \textbf{0.270} & \textbf{0.264} && \textbf{46.183} & \textbf{29.675} &\textbf{36.848}\\
\bottomrule
\end{tabular}
\end{table*}

\begin{figure*}[t]
    \centering
    \setlength{\abovecaptionskip}{0.cm}
    \includegraphics[width=1\textwidth]{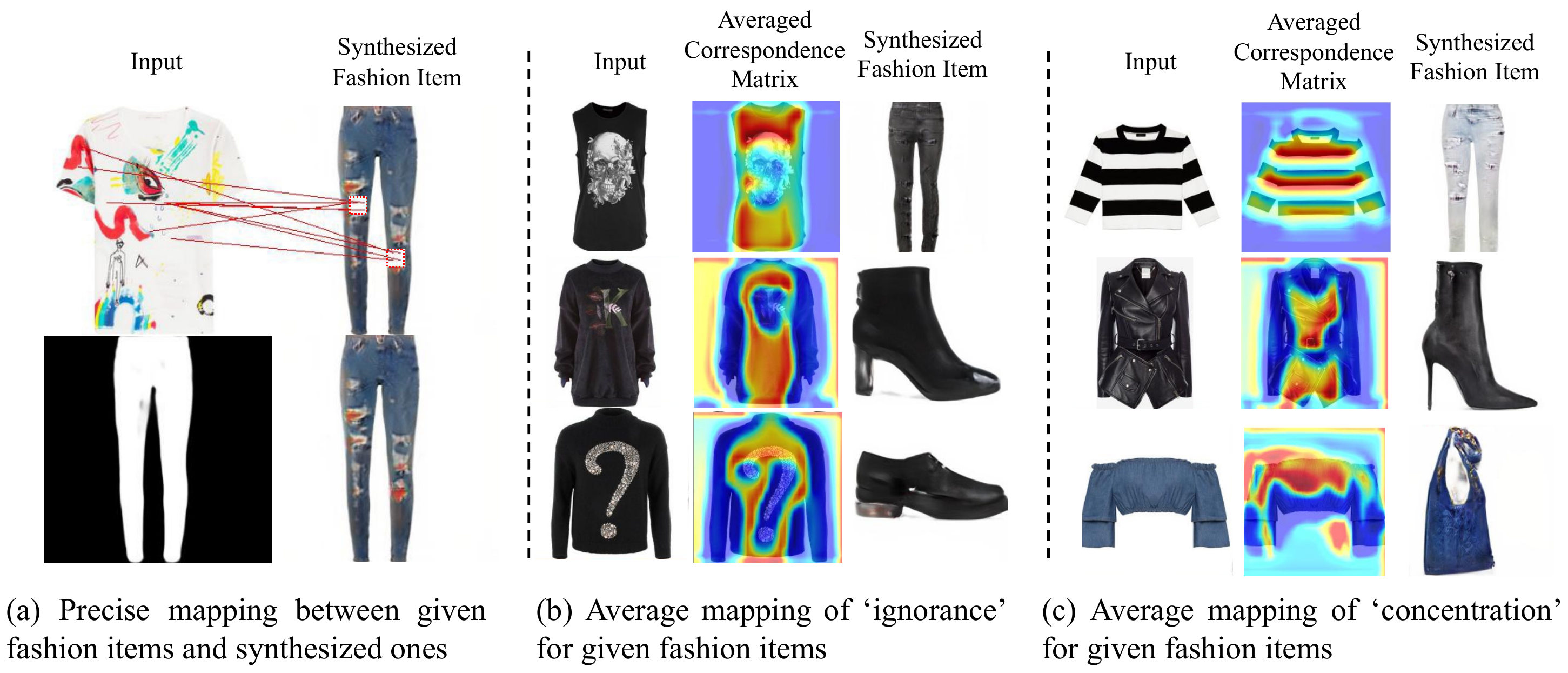}
    \caption{Illustration of explanation of our semantic alignment module for outfit generation.}
    \label{explanation}
\end{figure*}

We also compared our models with baseline methods with respect to the authenticity measurement, i.e., the FID. We evaluated the FID for each category of synthesized and target fashion items. For this metric, OutfitGAN reduces approximate 5.512, 25.724, and 6.649 for the generation of bag, lower clothing and shoes, respectively, in comparison with the second-best method. Our models with synthesized reference masks, OutfitGAN-P/OutfitGAN-R, reduce approximate 18.154, 9.825, and 7.088/ (10.886, 3.621, and 4.345) values of FID for the generation of bag, lower clothing and shoes compared with the second-best method. From the synthesized results in Fig. \ref{samples}, we can see that the images produced by our models have higher authenticity based on human perceptual observations. In particular, the Pix2Pix method sometimes produces spots on the borderline between the bag and lower clothing, and its synthesized images are not well contoured. The synthesis results from Pix2PixHD, MUNIT and DRIT++ are blurred, and the MUNIT method exhibits mode collapse in the synthesis of bags. CycleGAN always translates an upper clothing image into an outfit of compatible items while maintaining very similar styles, even for textual logos or lines. This can be attributed to the cycle reconstruction loss in CycleGAN. Of the methods compared here, StarGAN-v2 produces the best fashion items for lowers. Our OutfitGAN is able to synthesize the most visually plausible results based on real reference masks. Using synthesized masks, our OutfitGAN-P can also synthesize plausible results. With respect to the compatibility measurement for the synthesized outfits, the results for the FCTS for our OutfitGAN suggest that a generator supervised by a CCM can produce synthetic outfits with a superior degree of matching in comparison to other baselines. CycleGAN also produced promising results for the FCTS, as shown in Table \ref{ssim_lpips_fid_table}; however, the outfits synthesized by CycleGAN are based on styles that are extremely similar to those of the input upper clothing images, as can be observed from Fig. \ref{samples}. The outfits synthesized by CycleGAN therefore did not achieve a high compatibility score from a human perspective, due to the lack of difference in style from the given fashion items.

\subsection{Ablation Study}

In this subsection, two sets of experiments are carried out to validate the effectiveness of the SAM and the CCM, which are the main components of OutfitGAN. 

\textbf{Effectiveness of the SAM:} In order to investigate the effectiveness of the SAM, we validated it from two perspectives. Firstly, we trained our OutfitGAN without the SAM. In Table \ref{ssim_lpips_fid_zero_channel_table}, `OutfitGAN w/o SAM' means that we concatenated a reference mask with only the feature from the  $i$-th encoder $\mathrm{Enc}_i$ and fed the concatenated feature into the $i$-th decoder $\mathrm{Dec}_i$. The results show that the OutfitGAN model with the SAM consistently outperformed the model without the SAM in terms of the SSIM, LPIPS, and FID. This indicates that the SAM in our original framework was able to learn a correspondence relationship between a given fashion item and the targeted collocation items, allowing the visual similarity and authenticity to be significantly improved. To further examine the impacts of the SAM, we elaborate the explanation of the correspondence  $M_i^{corr}$ for the $i$-th synthesized fashion item during the generation process in Fig. \ref{explanation}. As shown in Fig. \ref{explanation}(a), there is a selected mapping relationship between the extant upper clothing and the synthesized lower clothing. For clarity, the corresponding four highest semantic regions for the areas of each white block in the synthesized lower images are annotated in Fig. \ref{explanation}(a). We can see that the red regions of synthesized images are always from the red patches or other salient patches of the extant upper clothing. This suggests that the SAM captures the translation correspondence relationship. In addition to the precise mapping between the given fashion items and the synthesized ones, we also average the correspondence matrix for visualization. It can be seen that the SAM cognitively processes the specific patterns of the given fashion items to some extent, as shown in Figs. \ref{explanation}(b) and (c). We can divide the mapping relationship for the given fashion items into two types: `ignorance' and `concentration'. Fig. \ref{explanation}(b) shows that the SAM can overlook some specific patterns for ignorance in given fashion items. Fig. \ref{explanation}(c) shows that the SAM concentrates more on certain patterns for outfit generation, and particularly on black-and-white lines rather than the other patterns. The above analysis of the SAM indicates that it is able to learn the correspondence relationships between the given fashion items and the synthesized ones.

\begin{table}[t]
\centering
\caption{Results of OutfitGAN in terms of collocation classification module on FCTS}
\label{cmp_nocmp_table}
\begin{tabular}{p{3.2cm} P{1.0cm}}
\toprule
Method & FCTS($\uparrow$)\\
 \hline
 OutfitGAN w/o CCM & 67.5\%\\
 OutfitGAN  w/ CCM & \textbf{87.1}\%\\
\bottomrule
\end{tabular}
\end{table}

\begin{figure*}[t]
    \centering
    \setlength{\abovecaptionskip}{0.cm}
    \includegraphics[width=0.95\textwidth, height=0.2714\textwidth]{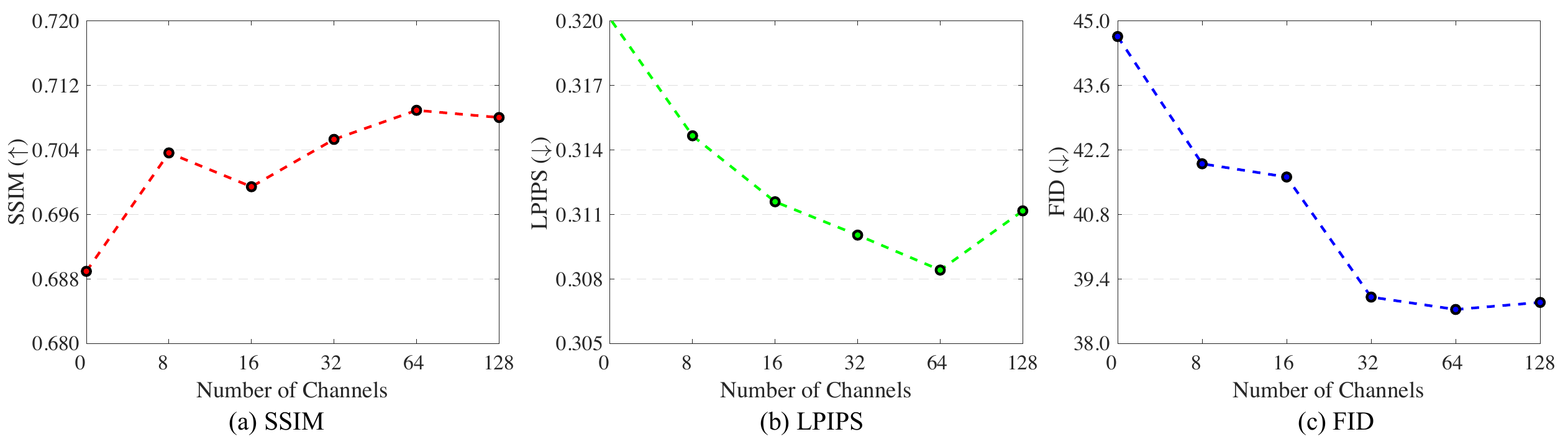}
    \caption{Similarity of synthesis measurements over all categories with the number of feature channels for semantic alignment module.}
    \label{channels}
\end{figure*}

\begin{figure}[t]
    \centering
    \setlength{\abovecaptionskip}{0.cm}
    \includegraphics[width=0.505\textwidth, height=0.29695423\textwidth]{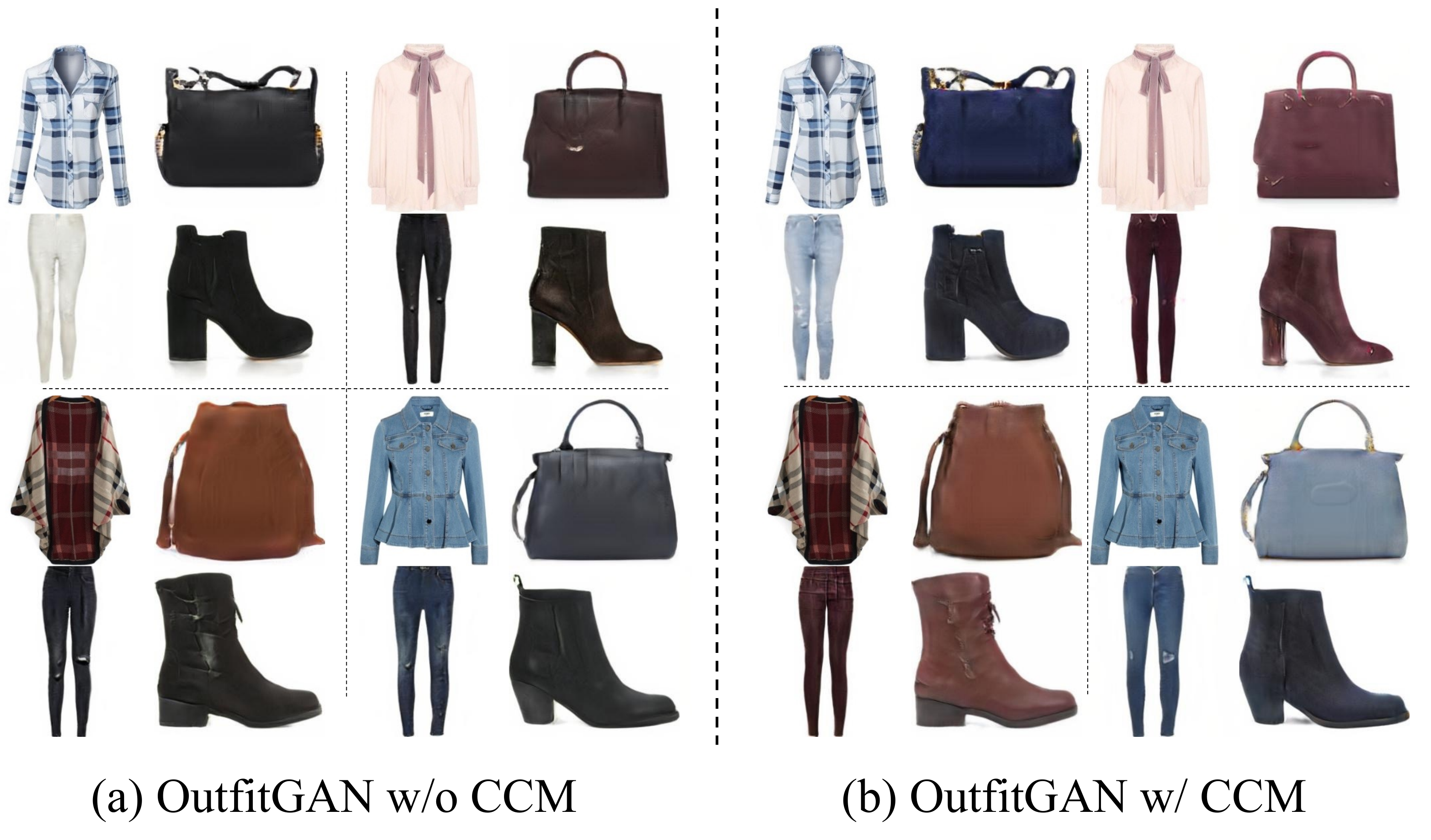}
    \caption{Synthesized samples of OutfitGAN with respect to collocation classification module.}
    \label{ccm0_1}
\end{figure}

\textbf{Effectiveness of the CCM:} We also explore the impact of the CCM in OutfitGAN. Specifically, we examine its effect on the FCTS in terms of visual compatibility. A comparison of the results is given in Table \ref{cmp_nocmp_table}, where `OutfitGAN w/ CCM' and `OutfitGAN w/o CCM' denote models with and without collocation classification, respectively. We can see that the model without the CCM gives a significant decrease in the FCTS, from 87.1\% to 67.5\%, thus demonstrating that the CCM markedly improves the compatibility of synthesized outfits. In addition, Fig. \ref{ccm0_1} shows that OutfitGAN with the CCM synthesizes more compatible outfits with more harmonious styles than the model without the CCM. The collocation module enhances the frequency of co-occurrence of compatible elements or style for compatible outfits. These quantitative and qualitative results suggest that the CCM can effectively improve the compatibility of the synthesized outfits.

\subsection{Parametric Study}
The main hyperparameters used in OutfitGAN are the number of feature channels for the SAM and the coefficients in the training losses.

\textbf{Number of feature channels for the SAM.} We first investigate the influence of the number of feature channels on the results of outfit generation. We set the number of feature channels $c$ to $[0, 8, 16, 32, 64, 128]$ in OutfitGAN. The results for the SSIM, LPIPS and FID over all categories are illustrated in Fig. \ref{channels}. In particular, the use of zero channel means that we concatenate only the reference mask with the feature extracted by the $i$-th encoder $\mathrm{Enc}_i$ in OutfitGAN. From Fig. \ref{channels}, we can see that an increase in the number of feature channels within the range $[0, 64]$ generally increases the performance of OutfitGAN in terms of the SSIM, LPIPS and FID. When the number of feature channels is increased beyond 64, the performance of OutfitGAN may become slightly worse. We ascribe this to the fact that the outfit generation process requires a much larger exploration space when the number of feature channels becomes large. In our experiments, setting the parameter $c$ to 64 was sufficient to deliver satisfactory results. 

\begin{figure}[t]
    \centering
    \setlength{\abovecaptionskip}{0.cm}
    \includegraphics[width=0.48\textwidth, height=0.514\textwidth]{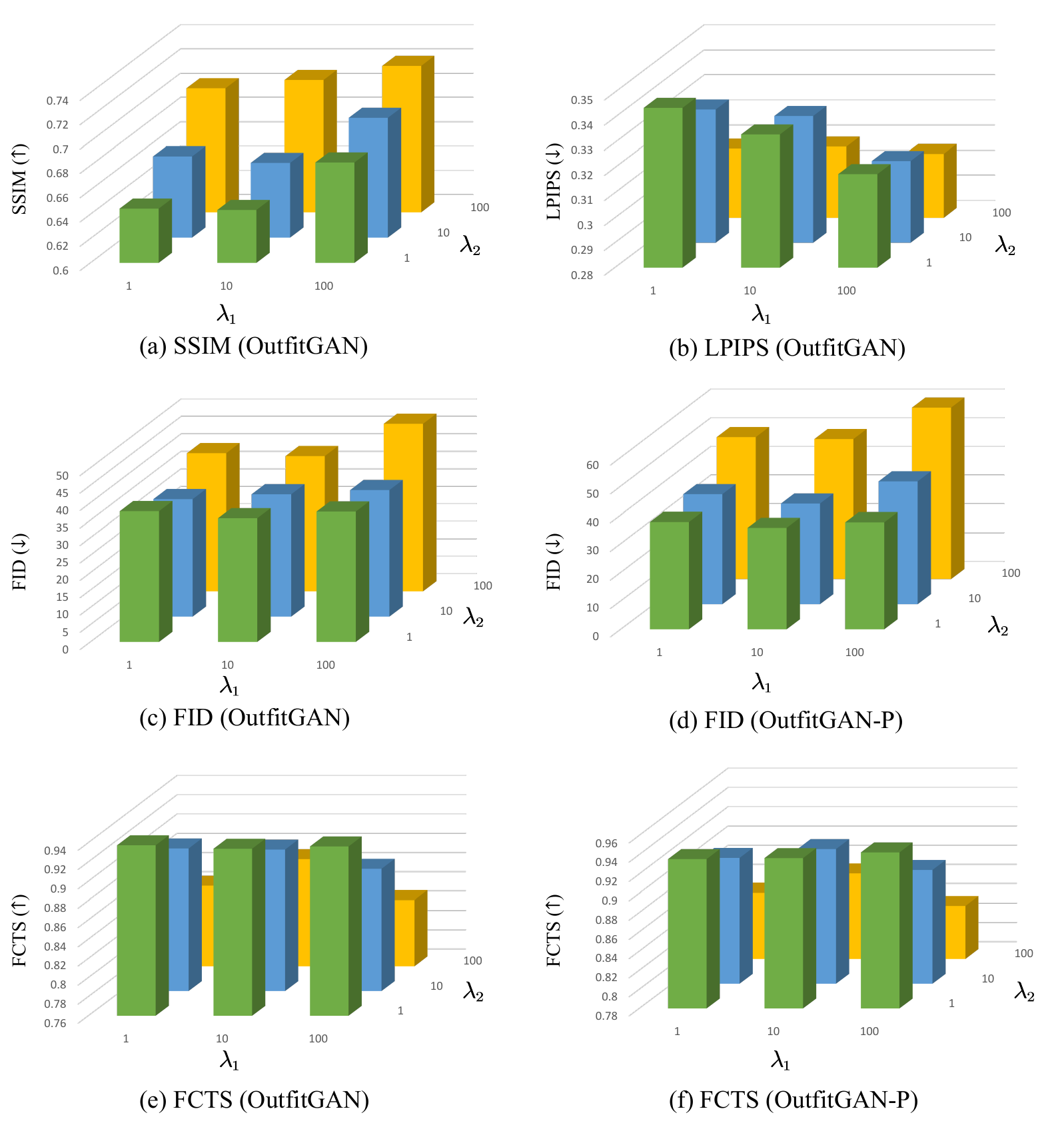}
    \caption{Results of OutfitGAN and OutfitGAN-P against different settings of weight coefficients of training losses.}
    \label{all_lambdas}
\end{figure}

\textbf{Coefficients of the training loss.} To further investigate the impacts of the coefficients used in weighting the training losses, we present the results from OutfitGAN with different weight parameters $\lambda _1$  and $\lambda _2$ for $\mathcal{L} _1$  and  $\mathcal{L} _{per}$, respectively, which are the reconstruction losses at the pixel-level and feature-level, as shown in Eqs. (\ref{l_1}) and (\ref{l_per}). The results are illustrated in Figs. \ref{all_lambdas} (a)-(f), in which the coefficients $\lambda _1$  and $\lambda _2$ are varied in the range $[1,10,100]$. Figs. \ref{all_lambdas} (a)-(c) show the results from OutfitGAN with different coefficients in terms of the metrics SSIM, LPIPS and FID; Fig. \ref{all_lambdas}(d) shows the FID results for OutfitGAN with Pix2Pix mask generation (i.e., OutfitGAN-P) for different coefficients; and Figs. \ref{all_lambdas}(e) and (f) show the FCTS results for OutfitGAN and OutfitGAN-P, respectively. From Fig. \ref{all_lambdas}, we observe that the parameters $\lambda_1$  and $\lambda_2$ have a significant impact on the similarity measurements, and an increase in these values can strongly improve the similarity between the synthesized fashion items and the target ones. However, $\lambda_1$  and $\lambda_2$ produce the opposite impact on the authenticity and compatibility of the synthesized fashion items. This can be ascribed to the fact that an increase in these parameters weakens the influence of the discriminator and the CCM in OutfitGAN. In our implementation, the selection of these coefficients was made based on a tradeoff between the similarity and authenticity measurements. The results show that settings of $\lambda _1=100$ and $\lambda _2=10$ for OutfitGAN and $\lambda _1=100$ and $\lambda _2=10$ for OutfitGAN-P gave the best synthesized images in terms of their similarity and authenticity. For simplicity, the coefficient settings for OutfitGAN-R were the same as those for OutfitGAN-P in our experiments.

\subsection{Study on Different Sequences of Fashion Items}

\begin{figure}[t]
	\centering
	\setlength{\abovecaptionskip}{0.cm}
	\includegraphics[width=0.49\textwidth, height=0.3047561\textwidth]{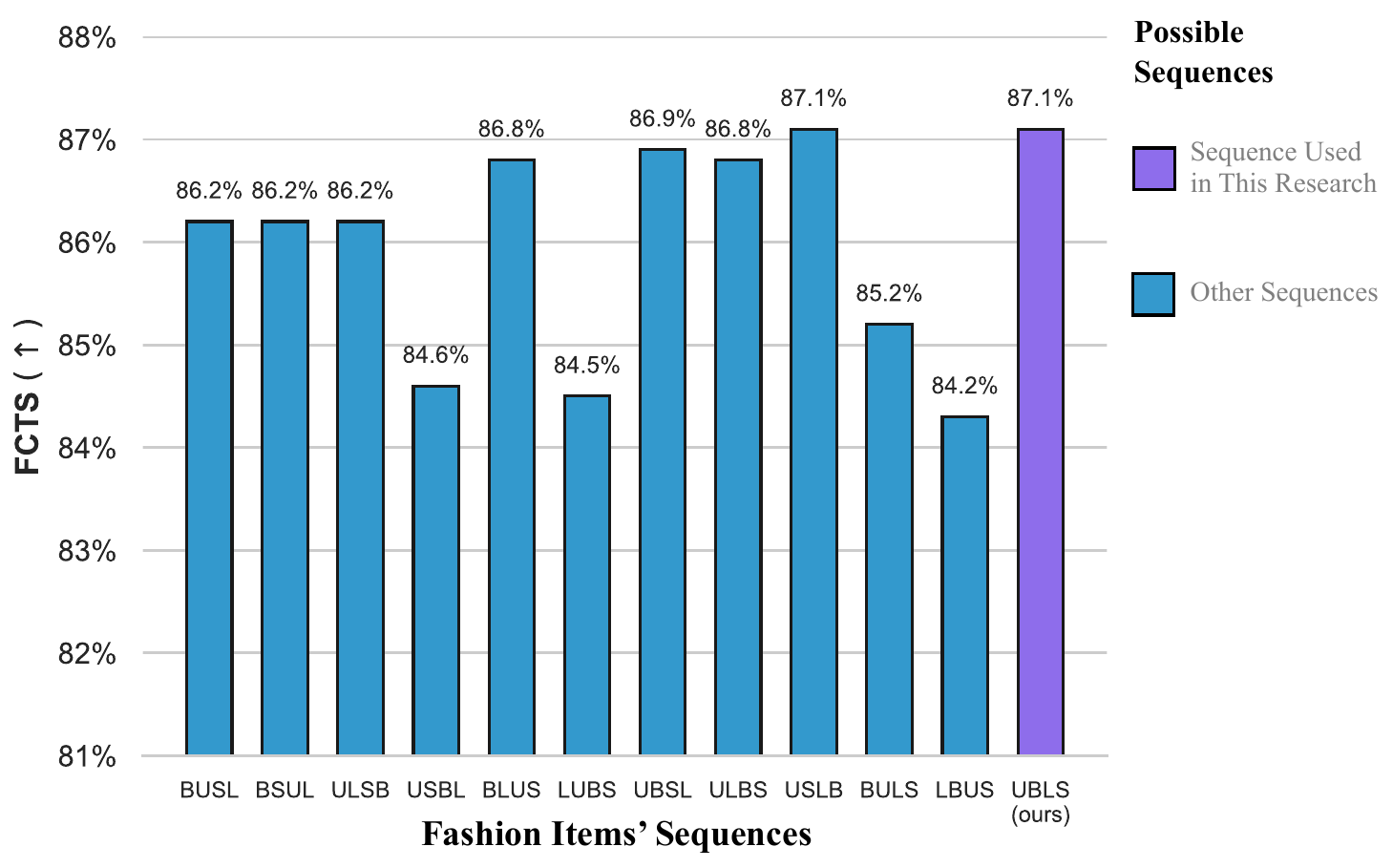}
	\caption{Fashion compatibility measurements against different settings of possible fashion items' sequences (here, each item in the abscissa represents a possible sequence, where `UBLS' represents an order of [upper, bag, lower, shoes], and other items have similar definitions).}
	\label{pos_seq}
\end{figure}

As previously stated, the fashion compatibility task can be addressed with a sequence model which is motivated by the human observation perspective \cite{han2017learning}. However, the sequence of the fashion items in an outfit has many possible arrangements. For an outfit with $N$ fashion items, it has $N!$ possible orders to model the fashion compatibility, where $N!$ denotes factorial $N$. In this section, we further investigate all possible sequences under our problem settings on the performance of OutfitGAN in terms of the fashion compatibility metric, FCTS. Considering that the used collocation classification module is based on Bi-LSTM, here we only have $\frac{N!}{2}$ possible unique orders in our task. We implemented different variants of OutfitGAN with all possible orders which were trained to validate the effectiveness of our pre-defined order, i.e., [upper, bag, lower, shoes]. Additional eleven versions of OutfitGAN were carried out in total. As shown in Fig. \ref{pos_seq}, we observe that our pre-defined order used in Section \ref{dataset_construct} obtains the best fashion compatibility in comparison to other possible orders, despite that `USLB' and `UBLS' have the same FCTS values (see Fig. \ref{pos_seq}). Moreover, most models with other orders show relatively decent performance on fashion compatibility. This may be ascribed to the fact that there only exist four fashion items in an outfit in our current research and Bi-LSTM may have sufficient ability to build the compatibility relation among fashion items in the same outfit even if we provide an arbitrary order.

\subsection{Limitation}
Although the proposed method achieves state-of-the-art performance in outfit generation, OutfitGAN still has certain limitations at the current stage. Firstly, an outfit includes $N$ fashion items, where $N=4$ in our implementation. During the process of our dataset construction, we crawled outfits which are composed by fashion experts from Polyvore.com. To cover as many fashion items as possible, we define our outfit generation on four commonly used items by women – upper, bag, lower, and shoes. It is possible to build a large-scale dataset with more kinds of fashion items when more relevant fashion compatibility-related resources are available in the future. Secondly, for an outfit with $N$  fashion items, OutfitGAN needs ($N-1$) item generators to synthesize the complementary fashion items based on the given item. The number of item generators increases with the number of fashion items, indicating that the computational complexity of OutfitGAN is $O(N)$. Actually, even if a shared item generator is used for synthesizing all kinds of fashion items, the model needs ($N-1$) feedforward times for synthesizing ($N-1$) fashion items. Moreover, once the outfit generator is trained, an arbitrary number of item generators can be selected for synthesizing desired fashion items. It is worth noting that every item generator is able to be used separately for synthesizing its targeted fashion item. For synthesizing multiple fashion items with lightweight models in more real-life applications, we leave this for future work.

\section{Conclusion}
\label{cnc}
This paper has presented an outfit generation framework with the aim of synthesizing photo-realistic fashion items that are compatible with a given item. In particular, in order to exploit the harmonious elements and styles shared in a compatible outfit, OutfitGAN uses a mask-guided strategy for image synthesis which can overcome the issue of spatial misalignment that arises in general image-to-image translation tasks. OutfitGAN consists of an outfit generator, an outfit discriminator and a CCM. An SAM is adopted to capture the mapping relationships between the extant fashion items and the synthesized ones, in order to improve the quality of fashion synthesis. A CCM is developed to improve the compatibility of the synthesized outfits. To evaluate the effectiveness of the proposed model, we constructed a large-scale dataset that consists of 20,000 outfits. Extensive experimental results show that our method can achieve state-of-the-art performance on the task of outfit generation and outperforms other methods. In the future, we plan to concentrate on synthesizing outfits with finer detail, and to use other reference information such as textual descriptions in a multi-modal manner to guide the process of outfit generation.

\appendices


%
%

\ifCLASSOPTIONcaptionsoff
  \newpage
\fi



%

\bibliographystyle{IEEEtran}
\bibliography{ref}


%







\end{document}